\documentclass[a4paper,10.5pt]{elsarticle}

\usepackage{geometry}
\usepackage{makecell}
\usepackage{extarrows}
\usepackage{subfigure}
\usepackage{mathrsfs}
\usepackage{lineno}
\usepackage{ulem}
\usepackage{graphicx}
\usepackage{subfigure}
\usepackage{makecell}
\usepackage{fancyhdr}
\usepackage{amssymb}
\usepackage{multicol}
\usepackage{amsmath}
\usepackage[figuresright]{rotating}

\usepackage{hyperref}
\usepackage{cleveref}
\usepackage[linesnumbered,ruled]{algorithm2e}
\usepackage{notoccite}

\usepackage{amsthm}
\theoremstyle{definition}

\newtheorem{thm}{Theorem}[section] %
\newtheorem{defn}[thm]{Definition} %
\newtheorem{pro}[thm]{Proposition} %
\newtheorem{exam}[thm]{Example}%

\newtheorem{rem}[thm]{Remark}

\begin{document}

\begin{frontmatter}

\title{Foundational theories of hesitant fuzzy sets and families of hesitant fuzzy sets}

\author{Shizhan Lu$^{1}$, Zeshui Xu$^{2}$, Zhu Fu$^{3}$, Longsheng Cheng$^{4*}$, Tongbin Yang$^{1}$}% , Longsheng Cheng$^1$, Haiyan Xu$^2$, Rashid Mehmood$^3$, Yu Han$^2$}

\address{\small  $^1$School of Management, Jiangsu University, Zhengjiang 212013, China\\% (lushizhan20140910@126.com) \\

$^2$Business School, Sichuan University, Chengdu, 610064, China\\% (xuzeshui@263.net)\\

$^3$School of Economics and Management, Jiangsu University of Science and Technology, Zhenjiang, 212100, China\\% (fuzhu886@163.com)\\

$^4$School of Economics and Management, Nanjing University of Science and Technology, Nanjing 210094, China\\

%\small  $^2$College of Economics and Management, Nanjing University of Aeronautics and Astronautics, Nanjing 211106,  China \\

%\small  $^3$Department of Software Engineering, University of Kotli, AJ$\&$K 11100, Pakistan. \\
}

\cortext[t1]{Corresponding author: Longsheng Cheng (chenglongshengnj@163.com).}%The work was supported by the National Natural Science Foundations of China (71904043).}%E-mail:}% Zeshui Xu (xuzeshui@263.net)}

\begin{abstract}

Hesitant fuzzy sets find extensive application in specific scenarios involving uncertainty and hesitation.
In the context of set theory, the concept of inclusion relationship holds significant importance as a fundamental definition.
Consequently, as a type of sets, hesitant fuzzy sets necessitate a clear and explicit definition of the inclusion relationship.
Based on the discrete form of hesitant fuzzy membership degrees, this study proposes multiple types of inclusion relationships for hesitant fuzzy sets.
Subsequently, this paper introduces foundational propositions related to hesitant fuzzy sets, as well as propositions concerning families of hesitant fuzzy sets.

\end{abstract}


\end{frontmatter}

%\linenumbers

\begin{multicols}{2}
\section{Introduction}

The foundation of modern mathematics lies in set theory established by Cantor \cite{FP} (referred to as classical sets hereafter).
Moreover, modern branches of mathematics, including group theory \cite{EJB,DDJ}, topology \cite{SWA}, graph theory \cite{GHH}, among others, form the underpinnings of computer science.
Subsequent to the refinement of classical set theory, various types of sets emerged, encompassing fuzzy sets \cite{ZADE,WuXHJiangDa}, interval-valued fuzzy sets \cite{GOMB}, intuitionistic fuzzy sets \cite{AKT}, and hesitant fuzzy sets \cite{TV}, among others.
With the exception of hesitant fuzzy sets, several types of sets have established their foundational theoretical frameworks.
Notably, the definition of the inclusion relationship between two hesitant fuzzy sets, which is one of the fundamental definitions, lacks adequate clarity.
This article develops the foundational theories of hesitant fuzzy sets and hesitant fuzzy information systems, and introduces a multi-strength intelligent classifier for diagnosing the health states of complex systems.

In the case of the classical set \cite{FP}, the membership degree of an element is described as either 0\% or 100\%, and the element's relationship with the set is determined by whether it belongs to the set or not.
Nevertheless, the real world is fraught with uncertainties. To address such uncertain scenarios, Zadeh introduced the concept of fuzzy sets \cite{ZADE}, where the membership degree of an element in a set is a value ranging from 0 to 1.
The membership degree of a fuzzy set denotes the probability of an element belonging to the set \cite{YSB}, thereby extending beyond the binary values of 0\% and 100\%.
Later, Gorzalczany \cite{GOMB} introduced the notion of interval-valued fuzzy sets, where the membership degree of an element in a set is represented by an interval included in $[0,1]$.
Interval-valued fuzzy sets can handle situations where the precise probability of an element belonging to a set cannot be determined, but an interval can be identified within which the probability must lie \cite{FKM}.
Additionally, Atanassov \cite{AKT} introduced the notion of  intuitionistic fuzzy sets, where the membership degree of an element in a set is represented by a bipartite array. Intuitionistic fuzzy sets can capture entities with dual characteristics, with one value in the bipartite array representing the positive aspect and the other value representing the negative aspect \cite{KRA, PLDY}.

Torra \cite{TV} introduced the concept of  hesitant fuzzy sets, where the membership degree of an element in a set is represented by a multidimensional array.
Hesitant fuzzy sets are designed to address situations of uncertainty and hesitation, which are prevalent in real-world scenarios.
For example, in scenarios involving data sampling for equipment monitoring and decision scoring by expert teams, the feedback often manifests as arrays with multiple diverse values, leading to indecisiveness.
Consequently, hesitant fuzzy sets with a membership degree represented by a discrete array are better suited for capturing the indecisiveness in decision-making compared to classical, fuzzy, interval-valued fuzzy, and intuitionistic fuzzy sets.
Hesitant fuzzy sets find extensive applications in various domains, including decision-making \cite{LLYX,GXLY},  attribute reduction \cite{zhang2024hesitant}, classification \cite{peng2024new}, linguistic perceptual \cite{AWNJE}, and forecasting \cite{PMNME}, among others.

Nevertheless, a sufficiently clear definition of the inclusion relationship between two hesitant fuzzy sets is currently lacking.
The definition of the inclusion relationship serves as a crucial foundation for sets, as it establishes equivalence between two sets when one is a subset of the other and vice versa.
Based on the discrete form of hesitant fuzzy membership degrees, this study introduces multiple types of inclusion relationships for hesitant fuzzy sets and subsequently presents foundational propositions concerning hesitant fuzzy sets and their families.
It is important to highlight that some foundational propositions applicable to classical sets do not hold in the case of hesitant fuzzy sets.

The remainder of this paper is organized as follows. Section 2 introduces various types of inclusion relationships among hesitant fuzzy sets, along with their foundational propositions, and shows whether certain rules that apply to classical sets also hold for hesitant fuzzy sets. In Section 3, we put forth propositions concerning families of hesitant fuzzy sets.
Section 4  provides a conclusion.

\section{Foundations of hesitant fuzzy sets}
%\section{Basic notions and results/ Preliminaries}

In this section, we provide a brief overview of fundamental concepts related to hesitant fuzzy sets and introduce various types of inclusion relationships that are applicable to the discrete form of hesitant fuzzy membership degrees.
Our focus is on examining whether certain rules that apply to classical sets also hold for hesitant fuzzy sets.
In further research on hesitant fuzzy sets, it is crucial for researchers to refrain from relying on the intuitions derived from classical sets and to be mindful of the rules that are valid in classical sets but not in hesitant fuzzy sets.

If two sets $A$ and $B$ are classical sets, $A\sqcap B$ and $A\sqcup B$ represent the intersection and union of $A$ and $B$, respectively. Furthermore, $A\sqsubset B$ represents that $A$ is a subset of the classical set $B$.
Let $U$ be a universal set and $E$ be a set of parameters.

\subsection{Reviews of fuzzy sets and hesitant fuzzy sets}

\begin{defn} \cite{ZADE} A fuzzy set $F$ on $U$ is a mapping $F:U\rightarrow [0,1]$.
\end{defn}

\begin{defn} \cite{ZADE} $F_1$ and $F_2$ are two fuzzy sets on $U$, $F_1$ is a subset of $F_2$ if $F_1(x)\leqslant F_2(x)$ for all $x\in U$, denoted as $F_1\subset F_2$.
\end{defn}

\begin{defn}\cite{XX} A hesitant fuzzy element is a non-empty, finite subset of $[0,1]$.
\end{defn}

\begin{defn}\cite{TV} A hesitant fuzzy set on $U$ is defined as a function  that when applied to $U$ returns a subset of $[0, 1]$.
\end{defn}

In the following, $HF(U)$ denotes the set of all hesitant fuzzy sets defined over $U$.

\begin{defn}\cite{TV} For each $x\in U$, and a hesitant fuzzy set $H$,  the lower bound and upper bound
of $H(x)$ are defined

lower bound $H^-(x)=inf\ H(x)$,

upper bound  $H^+(x)=sup\ H(x)$.
\end{defn}

\begin{defn}\cite{TV} Given two hesitant fuzzy sets represented by their membership functions
$H_1$ and $H_2$, their union and intersection are defined

union  $(H_1\cup H_2)(x)=\{h\in H_1(x)\sqcup H_2(x): h\geqslant sup(H_1^-(x),H_2^-(x))\}$,

intersection $(H_1\cap H_2)(x)=\{h\in H_1(x)\sqcup H_2(x): h\leqslant inf(H_1^+(x),H_2^+(x))\}$.
\end{defn}

\begin{defn} \cite{TV}  For $x\in U$ and $H\in HF(U)$, the complement of $H$ is denoted as $H^c$, where

$H^c(x)=\sqcup_{\gamma\in H(x)}\{1-\gamma\}$.

\end{defn}

\begin{thm}\label{thm8} \cite{TV,XX} The following statements hold for $A,B,C\in HF(U)$,

(1) $(A^c)^c= A$.

(2) $(A\cap B)^c= A^c\cup B^c$.

(3) $(A\cup B)^c= A^c\cap B^c$.

(4) $A\cap B= B\cap A$, $A\cup B= B\cup A$.

(5) $(A\cap B)\cap C= A\cap(B\cap C)$, $(A\cup B)\cup C= A\cup(B\cup C)$.
\end{thm}

Let $U=\{x,y\}$, $A(x)=\{0.6,0.5,0.3\}$ and $A(y)=\{0.5,0.3,0.2\}$. There are three common expression types for the hesitant fuzzy set $A$, shown as follows,
$$\begin{cases}
(type\ 1)\ \ A=\{\frac{\{0.6,0.5,0.3\}}{x},\frac{\{0.5,0.3,0.2\}}{y}\},\\
(type\ 2)\ \ A=\frac{\{0.6,0.5,0.3\}}{x}+\frac{\{0.5,0.3,0.2\}}{y},\\
(type\ 3)\ \ A=\{\langle x,(0.6,0.5,0.3)\rangle,\langle y,(0.5,0.3,0.2)\rangle\}.\\
\end{cases}$$
Furthermore,  $A=B$ means that $A(x)$ and $B(x)$ are  perfectly consistent for each $x\in U$.  For example,  $B=\frac{\{0.3,0.6,0.5\}}{x}+\frac{\{0.2,0.5,0.3\}}{y}$, then $B=A$;
$C=\frac{\{0.3,0.3,0.6,0.5\}}{x}+\frac{\{0.2,0.5,0.3\}}{y}$, then $C\neq A$.

\subsection{The proposed inclusion definitions of hesitant fuzzy sets and their relationships}

For two hesitant fuzzy sets $H_1$ and $H_2$, Babitha et al \cite{BJ} defined that $H_1$ is a hesitant fuzzy subset of $H_2$ if $H_1(x)\subset H_2(x)$ for all $x\in U$. However, the reference \cite{BJ} lacks a detailed description of   $H_1(x)\subset H_2(x)$.

For two fuzzy sets $F_1$ and $F_2$, we have $F_1\subset F_2\Leftrightarrow F_1\cup F_2=F_2$. Carlos et al \cite{JCVT} introduced a definition of inclusion relationship for hesitant fuzzy sets, i.e., $H_1\subset H_2$ $\Leftrightarrow$ $H_1\cup H_2=H_2$. This definition (in \cite{JCVT}) inherits the idea of fuzzy sets, which is convenient for scholars to refer to the existing theory of fuzzy sets to study hesitant fuzzy sets. However, we hold two different views on this definition (in \cite{JCVT}),  shown as (i) and (ii),

(i) The inclusion relationship of sets is also the ordering relationship of sets. A foundational and important proposition of inclusion relationship, $A\subset B$ and $B\subset A$ if and only if $A=B$, i.e., the equivalent sets contain each other. We assume that $H_1=H_2$ means that $H_1(x)$ and $H_2(x)$ are  perfectly consistent for each $x\in U$. By the description in \cite{JCVT}, $H_1(x)\subset H_2(x)\Leftrightarrow \{h:h\in H_2(x)\}=H_2(x)=(H_1\cup H_2)(x)=\{h\in H_1(x)\sqcup H_2(x):h\geqslant sup(H_1^-(x),H_2^-(x))\}$, i.e., $H_1(x)\subset H_2(x)$ implies $h<H_2^-(x)$ for all $h\in H_1(x)$. If $H_1(x)\subset H_2(x)$ and $H_2(x)\subset H_1(x)$, then a contradiction is produced, i.e.,  $H_1^+(x)<H_2^-(x)\leqslant H_2^+(x)<H_1^-(x)\leqslant H_1^+(x)$.

(ii) The definition of inclusion relationship in \cite{JCVT} is too specific to have a small range of applications. For example, let $H_1(x)=\{0.1,0.2,0.5\}$, $H_2(x)=\{0.6,0.7,0.9\}$ and $H_3(x)=\{0.1,0.5,0.8\}$, then $H_1(x)\subset H_2(x)$ is obvious. However, this definition cannot describe the relationship of $H_2(x)$ and $H_3(x)$ and is not applicable for many cases of hesitant fuzzy sets.

This study introduces various types of inclusion relations for hesitant fuzzy sets determined by the strength of the information contained in hesitant fuzzy membership degrees. Initially, a comprehensive example is provided to clarify Definition \ref{def2}.

\begin{exam}\label{examstart} Let $U=\{x_1,x_2,x_3,x_4,x_5,x_6\}$ be a set of decision-making schemes and $H$ be an expert team that consists of three experts. $H(U)=\frac{\{0.9,0.2\}}{x_1}+\frac{\{0.6,0.6,0.5\}}{x_2}+\frac{\{0.7,0.5,0.5\}}{x_3}+\frac{\{0.8,0.6,0.5\}}{x_4}+\frac{\{0.9,0.3,0.1\}}{x_5}+\frac{\{0.9,0.8,0.7\}}{x_6}$ are the estimated values of schemes provided by experts, in which the estimated values for $x_1$ are 0.9 and 0.2 that are obtained through the evaluations made by experts. One of the three experts fails to evaluate the scheme $x_1$.

(1) Here, $x_1$ has an estimated value of 0.9, which is greater than or equal to all the estimated values for $x_2$, it is possible that the scheme $x_1$ is better than the scheme $x_2$, denoted as $H(x_2)\subset_pH(x_1)$.

(2) Here, $0.55=mean[H(x_1)]<mean[H(x_2)]=0.567$, where $mean[\cdot]$ is the mean value operator. In comparing the mean values of estimated values, we find that the scheme $x_2$ is better than the scheme $x_1$, denoted as $H(x_1)\subset_mH(x_2)$.

(3) On the one hand, the best estimated value of $x_3$ is 0.7, which is greater than or equal to the best estimated value of $x_2$. On the other hand, the worst estimated value of $x_3$ is 0.5, which
is greater than or equal to the worst estimated value of $x_2$. To compare the respective best and worst cases of schemes $x_2$ and $x_3$, it is acceptable that the scheme $x_3$ is better than the scheme $x_2$, denoted as $H(x_2)\subset_aH(x_3)$.

(4) To compare the estimated values for schemes $x_3$ and $x_4$ one by one ($0.7\leqslant 0.8$; $0.5\leqslant 0.6$; and $0.5\leqslant0.5$), it is strongly credible that the scheme $x_4$ is better than the scheme $x_3$, denoted as $H(x_3)\subset_sH(x_4)$.

(5) We can obtain $H(x_1)\subset_sH(x_5)$ after truncating the tail estimated value of $x_5$, i.e., deleting the estimated value 0.1 of $x_5$. This case is denoted as $H(x_1)\subset_{st}H(x_5)$ and is recorded briefly as $H(x_1)\subset_{t}H(x_5)$.

(6) The worst estimated value of $x_6$ is greater than or equal to the best estimated value of $x_3$. Thus, it is necessary
that the scheme $x_6$ is better than the scheme $x_3$, which is denoted as $H(x_3)\subset_nH(x_6)$.

Comparing the mean value of estimated values is a common approach for decision-making;  however, while doing so, some important information may be lost, such as the best and the worst estimated values.
\end{exam}

\begin{defn}\label{def2} Let $H_1$ and $H_2$ be two hesitant fuzzy sets on $U$. Several kinds of inclusion relationships of two hesitant fuzzy sets are defined as follows,

(1) If $H^+_1(x)\leqslant H^+_2(x)$, then $H_1(x)\subset_p H_2(x)$. If $H_1(x)\subset_p H_2(x)$ for all $x\in U$, then $H_1\subset_p H_2$. If $H_1\subset_p H_2$ and $H_2\subset_p H_1$, then $H_1=_p H_2$.

(2) If $H^+_1(x)\leqslant H^+_2(x)$ and $H^-_1(x)\leqslant H^-_2(x)$, then $H_1(x)\subset_a H_2(x)$. If $H_1(x)\subset_a H_2(x)$ for all $x\in U$, then $H_1\subset_a H_2$. If $H_1\subset_a H_2$ and $H_2\subset_a H_1$, then $H_1=_a H_2$.

(3) If $mean[H_1(x)]\leqslant mean[H_2(x)]$, then $H_1(x)\subset_m H_2(x)$. If $H_1(x)\subset_m H_2(x)$ for all $x\in U$, then $H_1\subset_m H_2$. If $H_1\subset_m H_2$ and $H_2\subset_m H_1$, then $H_1=_m H_2$.

(4) Let $H_1(x)=V=\{v_1,v_2,\cdots,v_k\}$  and $H_2(x)=W=\{w_1,w_2,\cdots,w_l\}$ be two descending sequences. If $k\geqslant l$ and $w_i\geqslant v_i$ for $1\leqslant i\leqslant l$, then $H_1(x)\subset_s H_2(x)$. If $H_1(x)\subset_s H_2(x)$ for all $x\in U$, then $H_1\subset_s H_2$. If $H_1\subset_s H_2$ and $H_2\subset_s H_1$, then $H_1=_s H_2$.

(5) Let $H_1(x)=V=\{v_1,v_2,\cdots,v_k\}$  and $H_2(x)=W=\{w_1,w_2,\cdots,w_l\}$ be two descending sequences. If  $k<l$ and $w_i\geqslant v_i$ for $1\leqslant i\leqslant k$, then $H_1(x)\subset_{t}H_2(x)$. If $H_1(x)\subset_{t} H_2(x)$ for all $x\in U$, then $H_1\subset_{t} H_2$. It is obvious that $H_1\subset_{t} H_2$ and $H_2\subset_{t} H_1$ cannot hold simultaneously.

(6) If $H_1^+(x)\leqslant H_2^-(x)$, then $H_1(x)\subset_n H_2(x)$. If $H_1(x)\subset_n H_2(x)$ for all $x\in U$, then $H_1\subset_n H_2$.  If $H_1\subset_n H_2$ and $H_2\subset_n H_1$, then $H_1=_n H_2$.

\end{defn}

{\bf Symbols.}
(1)  Let $W=\{w_1,w_2,\cdots,w_l\}$ and $W'=\{w_1',w_2',\cdots,w_l'\}$ be two descending sequences and $|W|=|W'|$.
If $w_1'\geqslant w_1$, $w_2'\geqslant w_2$,$\cdots$,$w_l'\geqslant w_l$, then the relationship of $W$ and $W'$ is denoted as $W\preccurlyeq W'$.

(2) Let $q$ be a positive integer. To take $q$ larger numbers of $W$ and construct a best $q$-subsequence of $W$, denoted as $W(q)$, i.e., $|W(q)|=q$ and $w\leqslant inf(W(q))$ for $w\in W-W(q)$. For example, let $W=\{0.9,0.8,0.7,0.65,0.6,0.5\}$, when $q=2$, $W(q)=\{0.9,0.8\}$; when $q=3$, $W(q)=\{0.9,0.8,0.7\}$.

\begin{rem}\label{proadd} (1) If $W$ is a subsequence of $H_2(x)$, $H_1(x)\preccurlyeq W$ and $|H_1(x)|<|H_2(x)|$, then $H_1(x)\subset_t H_2(x)$, vice versa.

(2) If $W$ is a subsequence of $H_2(x)$, $H_1(x)\preccurlyeq W$ and $|H_1(x)|=|H_2(x)|$, then $H_1(x)\subset_s H_2(x)$.

(3) Let $q=inf\{|H_1(x)|,|H_2(x)|\}$, $H_1(x)(q)$ and $H_2(x)(q)$ be the best $q$-subsequences of $H_1(x)$ and $H_2(x)$, respectively.   Thereafter, $H_1(x)(q)\preccurlyeq H_2(x)(q)$ if and only if one of $H_1(x)\subset_s H_2(x)$ and $H_1(x)\subset_t H_2(x)$ holds.

(4) Let $K=H_1(x)\sqcap H_2(x)$, if $sup(H_1(x)-K)\leqslant inf(K)\leqslant sup(K)\leqslant inf(H_2(x)-K)$, then one of $H_1(x)\subset_s H_2(x)$ and $H_1(x)\subset_t H_2(x)$ holds.
\end{rem}

\begin{pro}\label{pro9ch} The following statements hold for $A,B\in HF(U)$,

(1) If $A\subset_a B$, then $A\subset_p B$.

(2) If $A\subset_s B$, then $A\subset_p B$.

(3) If $A\subset_s B$, then $A\subset_a B$.

(4) If $A\subset_s B$, then $A\subset_m B$.

(5) If $A\subset_t B$, then $A\subset_p B$.

(6) If  $A\subset_n B$, then $A\subset_p B$.

(7) If  $A\subset_n B$, then $A\subset_a B$.

(8) If  $A\subset_n B$, then $A\subset_m B$.

(9) If  $A\subset_n B$, then one of $A(x)\subset_s B(x)$ and $A(x)\subset_t B(x)$ holds for all $x\in U$.

\end{pro}

{\bf\slshape Proof} (1)-(8) are obvious.

(9) If $A\subset_n B$, then $A^+(x)\leqslant B^-(x)$ for all $x\in U$. If $|A(x)|\geqslant|B(x)|$, then $A(x)\subset_s B(x)$. If $|A(x)|<|B(x)|$, then $A(x)\subset_t B(x)$.$\blacksquare$

A brief summary for Proposition \ref{pro9ch} is shown as follows,
$$\begin{cases}
``\subset_n"\Rightarrow``\subset_s"\Rightarrow``\subset_a"\Rightarrow``\subset_p",  \\
``\subset_n"\Rightarrow``\subset_s"\Rightarrow``\subset_m",  \\
``\subset_n"\Rightarrow``\subset_t"\Rightarrow``\subset_p".  \\
\end{cases}$$

\subsection{Some propositions about the intersection and union of two hesitant fuzzy sets}

Let $\bar{A}$ and $\bar{B}$ be two classical sets, we know that $\bar{A}\sqcap \bar{B}\sqsubset \bar{A}$, $\bar{A}\sqcap \bar{B}\sqsubset \bar{B}$, $\bar{A}\sqsubset \bar{A}\sqcup \bar{B}$, $\bar{B}\sqsubset \bar{A}\sqcup \bar{B}$, and $\bar{A}\sqcap\bar{B}\sqsubset \bar{A}\sqcup \bar{B}$. However, these rules are not necessarily true for two hesitant fuzzy sets $A$ and $B$ (shown as Propositions \ref{pro2.9}-\ref{pro2.15}).

\begin{pro}\label{pro2.9} The following statements hold for $A_1,A_2\in HF(U)$,

(1) $A_1\cap A_2\subset_p A_i$ for $i=1,2$.

(2) $A_i\subset_p A_1\cup A_2$ for $i=1,2$.

(3) $A_1\cap A_2\subset_p A_1\cup A_2$.
\end{pro}

{\bf\slshape Proof} (1) and (2) are obvious.

(3) $inf\{A_1^+(x),A_2^+(x)\}\leqslant sup\{A_1^+(x),A_2^+(x)\}$.$\blacksquare$

\begin{pro} The following statements hold for $A_1,A_2\in HF(U)$,

(1) $A_1\cap A_2\subset_a A_i$ for $i=1,2$.

(2) $A_i\subset_a A_1\cup A_2$ for $i=1,2$.

(3) $A_1\cap A_2\subset_a A_1\cup A_2$.
\end{pro}

{\bf\slshape Proof} (1) and (2) are obvious.

(3) $inf\{A_1^+(x),A_2^+(x)\}\leqslant sup\{A_1^+(x),A_2^+(x)\}$ and $inf\{A_1^-(x),A_2^-(x)\}\leqslant sup\{A_1^-(x),A_2^-(x)\}$.$\blacksquare$

\begin{pro}\label{pro3} The following statements hold for $x\in U$ and $A,B\in HF(U)$,

(1) At least one of $(A\cap B)(x)\subset_m A(x)$ and $(A\cap B)(x)\subset_m B(x)$ holds.

(2)  At least one of $A(x)\subset_m (A\cup B)(x)$ and $B(x)\subset_m (A\cup B)(x)$ holds.

(3) $A\cap B\subset_m A\cup B$.

(4) If $A\subset_m B$, then $A\cap B\subset_m B$.

(5) If $A\subset_m B$, then $A\subset_m A\cup B$.
\end{pro}

{\bf\slshape Proof} For $x\in U$, denote $k_1=|A(x)|$, $k_2=|B(x)|$, $\bar{a}=mean[A(x)]$, and $\bar{b}=mean[B(x)]$, where $mean[\cdot]$ is the mean operator. Then $\sum\limits_{h\in A(x)}h+\sum\limits_{h\in B(x)}h=k_1\bar{a}+k_2\bar{b}$.

(1) For each $x\in U$, we can divide the relationships of $A^+(x)$ and $B^+(x)$ into two cases (c1) and (c2) as follows,

(c1) $A^+(x)=B^+(x)$.

$\sum\limits_{h\in (A\cap B)(x)}h=\sum\limits_{h\in A(x)}h+\sum\limits_{h\in B(x)}h=k_1\bar{a}+k_2\bar{b}\leqslant sup(\bar{a},\bar{b})(k_1+k_2)$. $mean[(A\cap B)(x)]=\sum\limits_{h\in (A\cap B)(x)}h/(k_1+k_2)\leqslant sup(\bar{a},\bar{b})(k_1+k_2)/(k_1+k_2)=sup(\bar{a},\bar{b})$. Then, one of  $(A\cap B)(x)\subset_m A(x)$ and $(A\cap B)(x)\subset_m B(x)$ holds.

(c2) $A^+(x)\neq B^+(x)$.

$A^+(x)\neq B^+(x)$, let $V=\{h:h\in A(x)\ or\ h\in B(x), h\not\in(A\cap B)(x)\}$ and $h'=inf(V)$, then $h'> inf(A^+(x),B^+(x))=sup\{h:h\in(A\cap B)(x)\}$.

Since $h'>sup\{h:h\in(A\cap B)(x)\}$ and $|(A\cap B)(x)|+|V|=k_1+k_2$, then $\sum\limits_{h\in (A\cap B)(x)}h/|(A\cap B)(x)|<(\sum\limits_{h\in (A\cap B)(x)}h+h'|V|)/(k_1+k_2)$. Since $h'=inf(V)$, then $(\sum\limits_{h\in (A\cap B)(x)}h+h'|V|)/(k_1+k_2)\leqslant (k_1\bar{a}+k_2\bar{b})/(k_1+k_2)$.

Then, we can obtain $mean[(A\cap B)(x)]=\sum\limits_{h\in (A\cap B)(x)}h/|(A\cap B)(x)|<(\sum\limits_{h\in (A\cap B)(x)}h+h'|V|)/(k_1+k_2)\leqslant (k_1\bar{a}+k_2\bar{b})/(k_1+k_2)\leqslant sup(\bar{a},\bar{b})(k_1+k_2)/(k_1+k_2)=sup(\bar{a},\bar{b})$. Then, one of $(A\cap B)(x)\subset_m A(x)$ and $(A\cap B)(x)\subset_m B(x)$ holds.

(2) For each $x\in U$, we can divide the relationships of $A^-(x)$ and $B^-(x)$ into two cases (c1) and (c2) as follows,

(c1) $A^-(x)=B^-(x)$.

$\sum\limits_{h\in (A\cup B)(x)}h=\sum\limits_{h\in A(x)}h+\sum\limits_{h\in B(x)}h=k_1\bar{a}+k_2\bar{b}\geqslant inf(\bar{a},\bar{b})(k_1+k_2)$.

$\sum\limits_{h\in (A\cup B)(x)}h/(k_1+k_2)\geqslant  inf(\bar{a},\bar{b})(k_1+k_2)/(k_1+k_2)= inf(\bar{a},\bar{b})$. Then, one of  $A(x)\subset_m (A\cup B)(x)$ and $B(x)\subset_m (A\cup B)(x)$ holds.

(c2) $A^-(x)\neq B^-(x)$.

$A^-(x)\neq B^-(x)$, let $V=\{h:h\in A(x)\ or\ h\in B(x), h\not\in(A\cup B)(x)\}$ and $h'=sup(V)$. $h'<sup(A^-(x),B^-(x))=inf\{h:h\in(A\cup B)(x)\}$.

Since $h'=sup(V)$ and $|(A\cup B)(x)|+|V|=k_1+k_2$, then $(\sum\limits_{h\in (A\cup B)(x)}h+h'|V|)/(k_1+k_2)\geqslant(k_1\bar{a}+k_2\bar{b})/(k_1+k_2)$.
Since $h'<inf\{h:h\in(A\cup B)(x)\}$, then $\sum\limits_{h\in (A\cup B)(x)}h/|(A\cup B)(x)|>(\sum\limits_{h\in (A\cup B)(x)}h+h'|V|)/(k_1+k_2)$.

Based on the results above, then  $mean[(A\cup B)(x)]=\sum\limits_{h\in (A\cup B)(x)}h/|(A\cup B)(x)|>(\sum\limits_{h\in (A\cup B)(x)}h+h'|V|)/(k_1+k_2)\geqslant(k_1\bar{a}+k_2\bar{b})/(k_1+k_2)\geqslant inf(\bar{a},\bar{b})(k_1+k_2)/(k_1+k_2)=inf(\bar{a},\bar{b})$. Then, one of $A(x)\subset_m (A\cup B)(x)$ and $B(x)\subset_m (A\cup B)(x)$ holds.

(3) For each $x\in U$, we can divide the relationships of $A^-(x)$, $A^+(x)$, $B^-(x)$ and $B^+(x)$ into four cases (c1), (c2), (c3) and (c4) as follows,

(c1) $A^-(x)=B^-(x)$ and $A^+(x)=B^+(x)$.

In this case, $(A\cap B)(x)= (A\cup B)(x)$, thus $mean[(A\cap B)(x)]=mean[(A\cup B)(x)]$, then $(A\cap B)(x)\subset_m (A\cup B)(x)$ is obvious.

(c2) $A^-(x)=B^-(x)$ and $A^+(x)\neq B^+(x)$.

In this case, $|(A\cap B)(x)|<k_1+k_2$ and $|(A\cup B)(x)|=k_1+k_2$. Let $V=\{h:h\in (A\cup B)(x), h\not\in (A\cap B)(x)\}$ and $h'=inf(V)$. $h'>inf(A^+(x),B^+(x))=sup\{h:h\in (A\cap B)(x)\}$.

Since $|(A\cap B)(x)|+|V|=|(A\cup B)(x)|=k_1+k_2$ and $h'>sup\{h:h\in (A\cap B)(x)\}$, then $\sum\limits_{h\in (A\cap B)(x)}h/|(A\cap B)(x)|<(\sum\limits_{h\in (A\cap B)(x)}h+h'|V|)/(k_1+k_2)$. Since $h'=inf(V)$, then $(\sum\limits_{h\in (A\cap B)(x)}h+h'|V|)/(k_1+k_2)\leqslant\sum\limits_{h\in (A\cup B)(x)}h/(k_1+k_2)$.

Based on the results above, then $mean[(A\cap B)(x)]=\sum\limits_{h\in (A\cap B)(x)}h/|(A\cap B)(x)|<(\sum\limits_{h\in (A\cap B)(x)}h+h'|V|)/(k_1+k_2)\leqslant\sum\limits_{h\in (A\cup B)(x)}h/(k_1+k_2)=mean[(A\cup B)(x)]$,
i.e., $(A\cap B)(x)\subset_m (A\cup B)(x)$.

(c3) $A^-(x)\neq B^-(x)$ and $A^+(x)=B^+(x)$.

In this case, $|(A\cap B)(x)|=k_1+k_2$ and $|(A\cup B)(x)|<k_1+k_2$. Let $V=\{h:h\in (A\cap B)(x), h\not\in (A\cup B)(x)\}$ and $h'=sup(V)$. $h'<sup(A^-(x),B^-(x))=inf\{h:h\in(A\cup B)(x)\}$.

Since $|(A\cup B)(x)|+|V|=|(A\cap B)(x)|=k_1+k_2$ and $h'<inf\{h:h\in(A\cup B)(x)\}$, then $\sum\limits_{h\in (A\cup B)(x)}h/|(A\cup B)(x)|>(\sum\limits_{h\in (A\cup B)(x)}h+h'|V|)/(k_1+k_2)$. Since $h'=sup(V)$, then $(\sum\limits_{h\in (A\cup B)(x)}h+h'|V|)/(k_1+k_2)\geqslant\sum\limits_{h\in (A\cap B)(x)}h/(k_1+k_2)$.

Based on the results above, then $mean[(A\cup B)(x)]=\sum\limits_{h\in (A\cup B)(x)}h/|(A\cup B)(x)|>(\sum\limits_{h\in (A\cup B)(x)}h+h'|V|)/(k_1+k_2)\geqslant\sum\limits_{h\in (A\cap B)(x)}h/(k_1+k_2)=mean[(A\cap B)(x)]$, i.e., $(A\cap B)(x)\subset_m (A\cup B)(x)$.

(c4) $A^-(x)\neq B^-(x)$ and $A^+(x)\neq B^+(x)$.

Based on the condition $A^-(x)\neq B^-(x)$, let $V=\{h:h\in(A\cap B)(x),h\not\in(A\cup B)(x)\}$ and $h'=sup(V)$, then $h'<sup(A^-(x), B^-(x))=inf\{h:h\in(A\cup B)(x)\}$, and $|V|+|(A\cup B)(x)|=k_1+k_2$.

Based on the results above, then $mean[(A\cup B)(x)]=\sum\limits_{h\in (A\cup B)(x)}h/|(A\cup B)(x)|>(\sum\limits_{h\in (A\cup B)(x)}h+h'|V|)/(k_1+k_2)\geqslant (k_1\bar{a}+k_2\bar{b})/(k_1+k_2)$.

Based on the condition $A^+(x)\neq B^+(x)$, let $W=\{h:h\in(A\cup B)(x),h\not\in(A\cap B)(x)\}$ and $h''=inf(W)$, then $h''> inf(A^+(x),B^+(x))=sup\{h:h\in(A\cap B)(x)\}$, and $|W|+|(A\cap B)(x)|=k_1+k_2$.

Based on the results above, then $mean[(A\cap B)(x)]=\sum\limits_{h\in (A\cap B)(x)}h/|(A\cap B)(x)|<(\sum\limits_{h\in (A\cap B)(x)}h+h''|W|)/(k_1+k_2)\leqslant(k_1\bar{a}+k_2\bar{b})/(k_1+k_2)<mean[(A\cup B)(x)]$, i.e., $(A\cap B)(x)\subset_m (A\cup B)(x)$.

To sum up, $A\cap B\subset_m A\cup B$.

(4) If $A\subset_m B$, by (1),  then $A\cap B\subset_m B$ is obtained.

(5) If $A\subset_m B$, by (2), then $A\subset_m A\cup B$ is obtained.$\blacksquare$

\begin{exam}\label{exam2.12}  Let $U=\{x,y,z\}$, $A=\frac{\{0.1,0.8\}}{x}+\frac{\{0.1,0.8\}}{y}+\frac{\{0.7,0.9\}}{z}$, $B=\frac{\{0.7,0.9\}}{x}+\frac{\{0.1,0.9\}}{y}+\frac{\{0.1,0.8\}}{z}$.

$(A\cap B)(x)=\{0.1,0.7,0.8\}$, $mean[(A\cap B)(x)]=0.53$. $mean[A(x)]=0.45$. $(A\cap B)(x)\not\subset_m A(x)$.

$(A\cup B)(y)=\{0.1,0.1,0.8,0.9\}$, $mean[(A\cup B)(y)]=0.475$. $mean[B(y)]=0.5$. $B(y)\not\subset_m(A\cup B)(y)$.

$(A\cap B)(y)\subset_m A(y)$ and $(A\cap B)(y)\subset_m B(y)$.

$A(x)\subset_m(A\cup B)(x)$ and $B(x)\subset_m(A\cup B)(x)$.
\end{exam}

For two random  hesitant fuzzy sets $A$ and $B$, Example \ref{exam2.12} shows that $(A\cap B)(x)\subset_m A(x)$ and $(A\cap B)(x)\subset_m B(x)$ do not hold simultaneously, $A(y)\subset_m(A\cup B)(y)$ and $B(y)\subset_m(A\cup B)(y)$ do not hold simultaneously.

$(A\cap B)(x)\subset_m B(x)$ and $(A\cap B)(x)\not\subset_m A(x)$. Furthermore, $(A\cap B)(z)\subset_m A(z)$ and $(A\cap B)(z)\not\subset_m B(z)$. It means that  (1) of Proposition \ref{pro3} cannot be written as that one of $A\cap B\subset_m A$ and $A\cap B\subset_m B$ holds. When we discuss the  case $``\subset_m"$ we need to note that
$$\begin{cases}
A_1\cap A_2\subset_m A_i\ \text{is uncertain for}\  i=1,2,  \\
A_i\subset_m A_1\cup A_2\ \text{is uncertain for}\  i=1,2.  \\
\end{cases}$$

\begin{pro}\label{pro1}  The following statements hold for $x\in U$ and $A_1,A_2\in HF(U)$,

(1) At least one of $A_i(x)\subset_s(A_1\cup A_2)(x)$ and $A_i(x)\subset_t(A_1\cup A_2)(x)$ holds for $i=1,2$.

(2) At least one of $(A_1\cap A_2)(x)\subset_s(A_1\cup A_2)(x)$ and $(A_1\cap A_2)(x)\subset_t(A_1\cup A_2)(x)$ holds.

\end{pro}

{\bf\slshape Proof} (1) We prove the case that one of $A_1(x)\subset_s(A_1\cup A_2)(x)$ and $A_1(x)\subset_t(A_1\cup A_2)(x)$ holds in follows.

For each $x\in U$, if  $A_1^-(x)\geqslant A_2^-(x)$, then $h\in(A_1\cup A_2)(x)$ for all $h\in A_1(x)$, i.e., $A_1(x)$ is a subsequence of $(A_1\cup A_2)(x)$. $A_1(x)\preccurlyeq A_1(x)$ and $|A_1(x)|\leqslant|(A_1\cup A_2)(x)|$, by Remark \ref{proadd} (1) and (2), then one of $A_1(x)\subset_s(A_1\cup A_2)(x)$ and $A_1(x)\subset_t(A_1\cup A_2)(x)$ holds.

If $A_1^-(x)< A_2^-(x)$, let $K=A_1(x)\sqcap (A_1\cup A_2)(x)$, then $sup(A_1(x)-K)<inf(K)\leqslant sup(K)<sup((A_1\cup A_2)(x)-K)$, by Remark \ref{proadd} (4), then one of $A_1(x)\subset_s(A_1\cup A_2)(x)$ and $A_1(x)\subset_t(A_1\cup A_2)(x)$ holds.

With the same manner, we can prove the case that one of $A_2(x)\subset_s(A_1\cup A_2)(x)$ and $A_2(x)\subset_t(A_1\cup A_2)(x)$ holds.

(2) Let $K=(A_1\cap A_2)(x)\sqcap (A_1\cup A_2)(x)$, then $sup((A_1\cap A_2)(x)-K)<inf(K)\leqslant sup(K)<inf((A_1\cup A_2)(x)-K)$, by Remark \ref{proadd} (4), one of $(A_1\cap A_2)(x)\subset_s(A_1\cup A_2)(x)$ and $(A_1\cap A_2)(x)\subset_t(A_1\cup A_2)(x)$ holds.$\blacksquare$\\

If two inclusion relationships $H_1(x)\subset_sH_2(x)$ and $H_1(x)\subset_tH_2(x)$ in Definition \ref{def2} are combined in $H_1(x)\subset_{sot}H_2(x)$ ($\subset_{sot}$ means $\subset_s$ or $\subset_t$), then  Proposition \ref{pro1} can be described as follows:

{\bf Proposition \ref{pro1}$'$} The following statements hold for $A_1,A_2\in HF(U)$,

(1) $A_i\subset_{sot}A_1\cup A_2$ for $i=1,2$.

(2) $A_1\cap A_2\subset_{sot}A_1\cup A_2$.

\begin{pro}\label{pro1nw}  The following statements hold for $x\in U$ and $A,B\in HF(U)$,

(1) If $A\subset_p B$, then at least one of $A(x)\subset_s(A\cap B)(x)$ and $A(x)\subset_t(A\cap B)(x)$ holds.

(2) If $A\subset_a B$, then at least one of $A(x)\subset_s(A\cap B)(x)$ and $A(x)\subset_t(A\cap B)(x)$ holds.

(3) If $A\subset_s B$, then at least one of $A(x)\subset_s(A\cap B)(x)$ and $A(x)\subset_t(A\cap B)(x)$ holds.

(4) If $A\subset_t B$, then at least one of $A(x)\subset_s(A\cap B)(x)$ and $A(x)\subset_t(A\cap B)(x)$ holds.

(5) If $A\subset_n B$, then at least one of $A(x)\subset_s(A\cap B)(x)$ and $A(x)\subset_t(A\cap B)(x)$ holds.

\end{pro}

{\bf\slshape Proof} (1) $A\subset_p B$ implies that $A^+(x)\leqslant B^+(x)$ for all $x\in U$. So we can deduce that if $h\in A(x)$ then $h\in(A\cap B)(x)$, i.e., $A(x)$ is a subsequence of $(A\cap B)(x)$.  $A(x)\preccurlyeq A(x)$ and $|A(x)|\leqslant |(A\cap B)(x)|$,  by Remark \ref{proadd} (1) and (2), then  one of $A(x)\subset_s(A\cap B)(x)$ and $A(x)\subset_t(A\cap B)(x)$ holds.

(2) $A\subset_a B$ implies $A\subset_p B$, by the result of (1), then (2) holds.

(3) $A\subset_s B$ implies $A\subset_p B$, by the result of (1), then (3) holds.

(4) $A\subset_t B$ implies $A\subset_p B$, by the result of (1), then (4) holds.

(5) $A\subset_n B$ implies $A\subset_p B$, by the result of (1), then (5) holds.$\blacksquare$\\

{\bf Proposition \ref{pro1nw}$'$} The following statements hold for $A,B,C\in HF(U)$,

(1) If $A\subset_p B$, then $A\subset_{sot}A\cap B$.

(2) If $A\subset_a B$, then $A\subset_{sot}A\cap B$.

(3) If $A\subset_{sot} B$, then $A\subset_{sot}A\cap B$.

(4) If $A\subset_n B$, then $A\subset_{sot}A\cap B$.

\begin{exam}  Let $U=\{x,y\}$, $A=\frac{\{0.1,0.2,0.5,0.6,0.9\}}{x}+\frac{\{0.1,0.7\}}{y}$, $B=\frac{\{0.05,0.3,0.4,0.7,0.8\}}{x}+\frac{\{0.8,0.9,0.9\}}{y}$.

$(A\cap B)(x)=\{0.05,0.1,0.2,0.3,0.4,0.5,0.6,0.7,0.8\}$, then $(A\cap B)(x)\not\subset_s A(x)$, $(A\cap B)(x)\not\subset_s B(x)$, $(A\cap B)(x)\not\subset_t A(x)$, and $(A\cap B)(x)\not\subset_t B(x)$.

$(A\cup B)(y)=\{0.8,0.9,0.9\}$, then $A(y)\subset_t (A\cup B)(y)$ and $B(y)\subset_s (A\cup B)(y)$.
\end{exam}

\begin{pro}\label{pro2.15}  The following statements hold for $A,B\in HF(U)$,

(1) If $A\subset_n B$, $A\cap B\subset_n B$.

(2) If $A\subset_n B$, $A\subset_n A\cup B$.

(3) If $A\subset_n B$, $A\cap B\subset_n A\cup B$.
\end{pro}

{\bf\slshape Proof} Since $A\subset_n B$, then $A^+(x)\leqslant B^-(x)$ for all $x\in U$, then (1), (2) and (3) are obvious.$\blacksquare$

\begin{exam} Let $U=\{x\}$, $A=\frac{\{0.1,0.3,0.5\}}{x}$, $B=\frac{\{0.2,0.4,0.6\}}{x}$.
\end{exam}

$(A\cap B)(x)=\{0.1,0.2,0.3,0.4,0.5\}$, $(A\cup B)(x)=\{0.2,0.3,0.4,0.5,0.6\}$. $A\cap B\not\subset_n A$, $A\cap B\not\subset_n B$, $A\not\subset_n A\cup B$, $B\not\subset_n A\cup B$, $A\cap B\not\subset_n A\cup B$.

\subsection{Some propositions about a pair of hesitant fuzzy sets (one being a subset of the other) and a random additional hesitant fuzzy set}

For three classical sets $\bar{A}$, $\bar{B}$ and $\bar{C}$, if $\bar{A}\sqsubset \bar{B}$, then $\bar{A}\sqsubset \bar{B}\sqcup \bar{C}$, $\bar{A}\sqcap \bar{C}\sqsubset \bar{B}\sqcap \bar{C}$  and $\bar{A}\sqcup \bar{C}\sqsubset \bar{B}\sqcup \bar{C}$.  These rules in the cases of three hesitant fuzzy sets $A$, $B$ and $C$ are investigated, shown as Proposition \ref{pro1nwa} and \ref{pro2}.

\begin{pro}\label{pro1nwa}  The following statements hold for $x\in U$ and $A,B,C\in HF(U)$,

(1) If $A\subset_p B$, then $A\subset_p B\cup C$.

(2) If $A\subset_a B$, then $A\subset_a B\cup C$.

(3) If $A\subset_s B$, then at least one of $A(x)\subset_s(B\cup C)(x)$ and $A(x)\subset_t(B\cup C)(x)$ holds.

(4) If $A\subset_t B$, then at least one of $A(x)\subset_s(B\cup C)(x)$ and $A(x)\subset_t(B\cup C)(x)$ holds.

(5) If $A\subset_n B$, then at least one of $A(x)\subset_s(B\cup C)(x)$ and $A(x)\subset_t(B\cup C)(x)$ holds.
\end{pro}

{\bf\slshape Proof} (1) and (2) are obvious.

(3) We prove this proposition after dividing the relationships of $h\in C(x)$ and the interval $[B^-(x),B^+(x)]$ into two cases  (i) and (ii) for each $x\in U$.

(i) $h\not\in[B^-(x),B^+(x)]$ for all $h\in C(x)$.

This case includes two sub-cases, $B^+(x)<C^-(x)$ and $C^+(x)<B^-(x)$.

If $B^+(x)<C^-(x)$, then $A(x)\subset_s B(x)\subset_n C(x)=(B\cup C)(x)$, then $A^+(x)\leqslant B^+(x)<C^-(x)$, i.e., $A(x)\subset_nC(x)=(B\cup C)(x)$. $A(x)\subset_n (B\cup C)(x)$, by Proposition \ref{pro9ch} (9), then one of $A(x)\subset_s (B\cup C)(x)$ and $A(x)\subset_t (B\cup C)(x)$ holds.

If $C^+(x)<B^-(x)$, then $A(x)\subset_s B(x)=(B\cup C)(x)$.

(ii) $h\in[B^-(x),B^+(x)]$ for some $h\in C(x)$.

This case includes two sub-cases, $C^-(x)\leqslant B^-(x)$ and $B^-(x)< C^-(x)$. Suppose $A(x)=\{o_1,o_2,\cdots,o_a\}=O$, $B(x)=\{v_1,v_2,\cdots,v_b\}=V$ and $C(x)=\{w_1,w_2,\cdots,w_c\}=W$, and $O$, $V$ and $W$ are descending sequences.

When $C^-(x)\leqslant B^-(x)$.

It is obvious that $h\in (B\cup C)(x)$ for all $h\in B(x)$. We can divide the elements of $A(x)$ into two descending sequences $\{o_1,o_2,\cdots,o_b\}$ and $\{o_{b+1},o_{b+2},\cdots,o_a\}$.

Since $A(x)\subset_s B(x)$, for $o_i\in\{o_1,o_2,\cdots,o_b\}$, we have $v_i\geqslant o_i$ and $v_i\in (B\cup C)(x)$, where $i=1,2,\cdots,b$.

For $o_j\in \{o_{b+1},o_{b+2},\cdots,o_a\}\sqsubset A(x)$ and  all $h\in(B\cup C)(x)$, we have $h\geqslant B^-(x)=v_b\geqslant o_b\geqslant o_{b+1}\geqslant o_j$, where $j=b+1,b+2,\cdots,a$.

Based on the illustrations of sub-case $C^-(x)\leqslant B^-(x)$, we have $A(x)(q)\preccurlyeq (B\cup C)(x)(q)$, where $q=inf\{|A(x)|,|(B\cup C)(x)|\}$, by Remark \ref{proadd} (3), one of $A(x)\subset_s (B\cup C)(x)$ and $A(x)\subset_t (B\cup C)(x)$ holds.

When $B^-(x)< C^-(x)$.

Suppose $v_{m-1}\geqslant w_c=C^-(x)$ and $v_m<w_c$. For all $h\in (B\cup C)(x)$, $h\geqslant w_c=C^-(x)$. We can divide the elements of $A(x)$ into two descending sequences $\{o_1,o_2,\cdots,o_{m-1}\}$ and $\{o_{m},o_{m+1},\cdots,o_a\}$.

For $o_i\in\{o_1,o_2,\cdots,o_{m-1}\}$, we have $v_i\geqslant o_i$ and $v_i\in (B\cup C)(x)$, where $i=1,2,\cdots,m-1$.

For $o_j\in \{o_{m},o_{m+1},\cdots,o_a\}\sqsubset A(x)$ and all $h\in(B\cup C)(x)$, we have $h\geqslant C^-(x)=w_c>v_m\geqslant o_m\geqslant o_j$, where $j=m,m+1,\cdots,a$.

Based on the illustrations of sub-case $B^-(x)< C^-(x)$, we have $A(x)(q)\preccurlyeq (B\cup C)(x)(q)$, where $q=inf\{|A(x)|,|(B\cup C)(x)|\}$, by Remark \ref{proadd} (3), one of $A(x)\subset_s (B\cup C)(x)$ and $A(x)\subset_t (B\cup C)(x)$ holds.

(4) We prove this proposition after dividing the relationships of $h\in C(x)$ and the interval $[B^-(x),B^+(x)]$ into two cases  (i) and (ii) for each $x\in U$.

(i) $h\not\in[B^-(x),B^+(x)]$ for all $h\in C(x)$.

This case includes two sub-cases, $B^+(x)<C^-(x)$ and $C^+(x)<B^-(x)$.

When $B^+(x)<C^-(x)$, we have $A(x)\subset_t B(x)\subset_n C(x)=(B\cup C)(x)$, then $A^+(x)\leqslant B^+(x)<C^-(x)$, i.e., $A(x)\subset_n C(x)=(B\cup C)(x)$. $A(x)\subset_n (B\cup C)(x)$, by Proposition \ref{pro9ch} (9),  then one of $A(x)\subset_s (B\cup C)(x)$ and $A(x)\subset_t (B\cup C)(x)$ holds.

When $C^+(x)<B^-(x)$, $A(x)\subset_t B(x)=(B\cup C)(x)$.

(ii) $h\in[B^-(x),B^+(x)]$ for some $h\in C(x)$.

This case includes two sub-cases, $C^-(x)\leqslant B^-(x)$ and $B^-(x)< C^-(x)$.

When $C^-(x)\leqslant B^-(x)$.

We have $h\in(B\cup C)(x)$ for all $h\in B(x)$, i.e., $B(x)\sqsubset (B\cup C)(x)$. Since $A(x)\subset_t B(x)$, by Remark \ref{proadd} (1), there is a subsequence of $B(x)$, denoted as $\overline{B(x)}$,  satisfying $A(x)\preccurlyeq\overline{B(x)}$. Since $B(x)\sqsubset (B\cup C)(x)$, then there is a subsequence of $(B\cup C)(x)$, denoted as $\overline{(B\cup C)(x)}$,  satisfying $A(x)\preccurlyeq\overline{(B\cup C)(x)}$. $|A(x)|<|B(x)|\leqslant|(B\cup C)(x)|$ is obvious. By Remark \ref{proadd} (1), $A(x)\subset_t (B\cup C)(x)$ holds.

When $B^-(x)< C^-(x)$.

Suppose $A(x)=\{o_1,o_2,\cdots,o_a\}=O$, $B(x)=\{v_1,v_2,\cdots,v_b\}=V$ and $C(x)=\{w_1,w_2,\cdots,w_c\}=W$, and $O$, $V$ and $W$ are descending sequences.

Suppose $v_{m-1}\geqslant w_c$ and $v_m< w_c=C^-(x)$. We can divide the elements of $A(x)$ into two descending sequences $\{o_1,o_2,\cdots,o_{m-1}\}$ and $\{o_m,o_{m+1},\cdots,o_a\}$.

Since $A(x)\subset_t B(x)$ and $v_{m-1}\geqslant w_c=C^-(x)$, then $\{o_1,o_2,\cdots,o_{m-1}\}\preccurlyeq\{v_1,v_2,\cdots,v_{m-1}\}$ for   $\{o_1,o_2,\cdots,o_{m-1}\}\sqsubset A(x)$  and $\{v_1,v_2,\cdots,v_{m-1}\}\sqsubset(B\cup C)(x)$.

Since $o_m\leqslant v_m< w_c=C^-(x)$, for  $o_j\in \{o_m,o_{m+1},\cdots,o_a\}\sqsubset A(x)$ and all $h\in (B\cup C)(x)$, it implies $h>o_j$, where $j=m,m+1,\cdots,a$.

Based on the illustrations of sub-case $B^-(x)< C^-(x)$, $A(x)(q)\preccurlyeq (B\cup C)(x)(q)$ is obtained, where $q=inf\{|A(x)|,|(B\cup C)(x)|\}$, by Remark \ref{proadd} (3), one of $A(x)\subset_s (B\cup C)(x)$ and $A(x)\subset_t (B\cup C)(x)$ holds.

(5) By Proposition \ref{pro9ch} (9), $A\subset_n B$ implies $A(x)\subset_sB(x)$ or $A(x)\subset_tB(x)$ for all $x\in U$. By the results of (3) and (4) of Proposition \ref{pro1nwa},   (5) holds.$\blacksquare$\\

(3) and (4) of Proposition \ref{pro1nwa} can be combined as (1) of Proposition \ref{pro1nwa}$'$.  (5) of Proposition \ref{pro1nwa} can be written into (2) of Proposition \ref{pro1nwa}$'$.

{\bf Proposition \ref{pro1nwa}$'$} The following statements hold for $A,B,C\in HF(U)$,

(1) If $A\subset_{sot} B$, then $A\subset_{sot}B\cup C$.

(2) If $A\subset_n B$, then $A\subset_{sot}B\cup C$.

\begin{pro}\label{pro2adnew} The following statements hold for $A,B,C\in HF(U)$,

(1) If $A\subset_p B$, $A\cap C\subset_p B\cap C$.

(2) If $A\subset_a B$, $A\cap C\subset_a B\cap C$.

(3) If $A\subset_s B$, $A\cap C\subset_a B\cap C$.

(4) If $A\subset_t B$, $A\cap C\subset_p B\cap C$.

(5) If $A\subset_n B$, $A\cap C\subset_a B\cap C$.

\end{pro}

{\bf\slshape Proof} (1)-(5) are obvious.

\begin{pro}\label{pro2} The following statements hold for $x\in U$ and $A,B,C\in HF(U)$,

(1) If $A\subset_p B$, $A\cup C\subset_p B\cup C$.

(2) If $A\subset_a B$, $A\cup C\subset_a B\cup C$.

(3) If $A\subset_s B$, $A\cup C\subset_a B\cup C$.

(4) If $A\subset_t B$, $A\cup C\subset_p B\cup C$.

(5)  If $A\subset_n B$, $A\cup C\subset_a B\cup C$.

(6) If $A\subset_s B$, then at least one of $(A\cup C)(x)\subset_s(B\cup C)(x)$ and $(A\cup C)(x)\subset_t(B\cup C)(x)$ holds.

(7) If $A\subset_t B$, then at least one of $(A\cup C)(x)\subset_s(B\cup C)(x)$ and $(A\cup C)(x)\subset_t(B\cup C)(x)$ holds.

(8)  If $A\subset_n B$, then at least one of $(A\cup C)(x)\subset_s(B\cup C)(x)$ and $(A\cup C)(x)\subset_t(B\cup C)(x)$ holds.

\end{pro}

{\bf\slshape Proof} (1)-(5) are obvious.

(6) For a random $x\in U$, suppose $A(x)=\{o_1,o_2,\cdots,o_a\}=O$, $B(x)=\{v_1,v_2,\cdots,v_b\}=V$ and $C(x)=\{w_1,w_2,\cdots,w_c\}=W$, and $O$, $V$ and $W$ are descending sequences. Since $A\subset_s B$, then $b\leqslant a$, and $o_i\leqslant v_i$ for $1\leqslant i\leqslant b$.

To sort and investigate the two sequences $W^1=\{o_1,o_2,\cdots,o_b\}\sqcup W$ and $W^2=\{v_1,v_2,\cdots,v_b\}\sqcup W$, it is obvious that $W^1\preccurlyeq W^2$.

$v_b=B^-(x)\leqslant(B\cup C)^-(x)$, then $(B\cup C)(x)$ is a subsequence of $W^2$. Let $q=inf\{|(A\cup C)(x)|,|(B\cup C)(x)|\}$, then $q\leqslant |W^1|=|W^2|$. $w^1_j\leqslant w^2_j$ for $1\leqslant j\leqslant q=inf\{|(A\cup C)(x)|,|(B\cup C)(x)|\}$, where $w^1_j\in W^1$ and $w^2_j\in W^2$.

By Remark \ref{proadd} (3), one of $(A\cup C)(x)\subset_s(B\cup C)(x)$ and $(A\cup C)(x)\subset_t(B\cup C)(x)$ holds.

(7) For a random $x\in U$, suppose $A(x)=\{o_1,o_2,\cdots,o_a\}=O$, $B(x)=\{v_1,v_2,\cdots,v_b\}=V$ and $C(x)=\{w_1,w_2,\cdots,w_c\}=W$, and $O$, $V$ and $W$ are descending sequences. Since $A\subset_t B$, then $a<b$, and $o_i\leqslant v_i$ for $1\leqslant i\leqslant a$.

To sort and investigate the two sequences $W^1=\{o_1,o_2,\cdots,o_a\}\sqcup W$ and $W^2=\{v_1,v_2,\cdots,v_a\}\sqcup W$, it is obvious that $W^1\preccurlyeq W^2$.

$o_a=A^-(x)\leqslant(A\cup C)^-(x)$, then $(A\cup C)(x)$ is a subsequence of $W^1$. Let $q=inf\{|(A\cup C)(x)|,|(B\cup C)(x)|\}$, then $q\leqslant |W^1|=|W^2|$. $w^1_j\leqslant w^2_j$ for $1\leqslant j\leqslant q=inf\{|(A\cup C)(x)|,|(B\cup C)(x)|\}$, where $w^1_j\in W^1$ and $w^2_j\in W^2$.

By Remark \ref{proadd} (3), one of $(A\cup C)(x)\subset_s(B\cup C)(x)$ and $(A\cup C)(x)\subset_t(B\cup C)(x)$ holds.

(8) By Proposition \ref{pro9ch} (9), $A\subset_n B$ implies $A(x)\subset_s B(x)$ or $A(x)\subset_t B(x)$ for $x\in U$. By the results of (6) and (7) of Proposition \ref{pro2}, then (8) holds.$\blacksquare$\\

(6) and (7) of Proposition \ref{pro2} can be combined as (1) of Proposition \ref{pro2}$'$.  (8) of Proposition \ref{pro2} can be written into (2) of Proposition \ref{pro2}$'$.

{\bf Proposition \ref{pro2}$'$} The following statements hold for $A,B,C\in HF(U)$,

(1) If $A\subset_{sot}B$,  then  $A\cup C\subset_{sot}B\cup C$.

(2) If $A\subset_n B$,  then  $A\cup C\subset_{sot}B\cup C$.

\begin{exam}\label{exam215}  Let $U=\{x_1,x_2,x_3,x_4,x_5,x_6,x_7,x_8\}$, $A=\frac{\{0.2,0.4\}}{x_1}+\frac{\{0.2,0.5\}}{x_2}+\frac{\{0.3,0.5\}}{x_3}+\frac{\{0.5,0.6\}}{x_4}+
\frac{\{0.6,0.7\}}{x_5}+\frac{\{0.3,0.4\}}{x_6}+\frac{\{0.3,0.4\}}{x_7}+\frac{\{0.3,0.4\}}{x_8}$, $B=\frac{\{0.1,0.1,0.5\}}{x_1}+\frac{\{0.1,0.8\}}{x_2}+\frac{\{0.4,0.41\}}{x_3}+\frac{\{0.6,0.7\}}{x_4}+
\frac{\{0.8,0.9\}}{x_5}+\frac{\{0.1,0.1,0.3,0.5\}}{x_6}+\frac{\{0.1,0.1,0.3,0.5\}}{x_7}+\frac{\{0.5,0.8\}}{x_8}$, $C=\frac{\{0.1,0.2\}}{x_1}+\frac{\{0.5\}}{x_2}+\frac{\{0.45,0.45\}}{x_3}+\frac{\{0.7,0.8\}}{x_4}+
\frac{\{0.1,0.7\}}{x_5}+\frac{\{0.3,0.4\}}{x_6}+\frac{\{0.1,0.4\}}{x_7}+\frac{\{0.8,0.9\}}{x_8}$.

(1) $A(x_1)\subset_p B(x_1)$, $(A\cap C)(x_1)=\{0.1,0.2,0.2\}$, $(B\cap C)(x_1)=\{0.1,0.1,0.1,0.2\}$.  $(A\cap C)(x_1)\not\subset_m (B\cap C)(x_1)$.
   $(A\cap C)(x_1)\not\subset_t (B\cap C)(x_1)$.

   $A(x_2)\subset_p B(x_2)$, $(A\cap C)(x_2)=\{0.2,0.5,0.5\}$, $(B\cap C)(x_2)=\{0.1,0.5\}$, $(A\cap C)(x_2)\not\subset_a(B\cap C)(x_2)$.

$A(x_1)\subset_p B(x_1)$, $(A\cup C)(x_1)=\{0.2,0.2,0.4\}$, $(B\cup C)(x_1)=\{0.1,0.1,0.1,0.2,0.5\}$. $(A\cup C)(x_1)\not\subset_m (B\cup C)(x_1)$. $(A\cup C)(x_1)\not\subset_a (B\cup C)(x_1)$.  $(A\cup C)(x_1)\not\subset_t (B\cup C)(x_1)$.

(2) $A(x_5)\subset_a B(x_5)$, $(A\cap C)(x_5)\not\subset_m (B\cap C)(x_5)$. $(A\cap C)(x_5)\not\subset_t (B\cap C)(x_5)$.

$A(x_8)\subset_a B(x_8)$, $(A\cup C)(x_8)\not\subset_m (B\cup C)(x_8)$.

$A(x_5)\subset_a B(x_5)$, $(A\cup C)(x_5)\not\subset_t (B\cup C)(x_5)$.

(3) $A(x_3)\subset_m B(x_3)$, $(A\cap C)(x_3)=\{0.3,0.45,0.45\}$, $(B\cap C)(x_3)=\{0.4,0.41\}$. $(A\cap C)(x_3)\not\subset_p (B\cap C)(x_3)$.

$A(x_2)\subset_m B(x_2)$,  $(A\cap C)(x_2)=\{0.2,0.5,0.5\}$, $(B\cap C)(x_2)=\{0.1,0.5\}$. $(A\cap C)(x_2)\not\subset_m (B\cap C)(x_2)$.

$A(x_3)\subset_m B(x_3)$, $(A\cup C)(x_3)=\{0.45,0.45,0.5\}$, $(B\cup C)(x_3)=\{0.45,0.45\}$. $(A\cup C)(x_3)\not\subset_p (B\cup C)(x_3)$.

$A(x_4)\subset_m B(x_4)$, $(A\cup C)(x_4)=\{0.7,0.8\}$, $(B\cup C)(x_4)=\{0.7,0.7,0.8\}$.  $(A\cup C)(x_4)\not\subset_m (B\cup C)(x_4)$.

(4) $A(x_5)\subset_s B(x_5)$, $(A\cap C)(x_5)=\{0.1,0.6,0.7,0.7\}$, $(B\cap C)(x_5)=\{0.1,0.7\}$. $(A\cap C)(x_5)\not\subset_m (B\cap C)(x_5)$.
 $(A\cap C)(x_5)\not\subset_t (B\cap C)(x_5)$.

$A(x_4)\subset_s B(x_4)$, $(A\cup C)(x_4)\not\subset_m (B\cup C)(x_4)$.

$A(x_5)\subset_s B(x_5)$, $(A\cup C)(x_5)\not\subset_t (B\cup C)(x_5)$.

(5) $A(x_6)\subset_t B(x_6)$, $(A\cap C)(x_6)=\{0.3,0.3,0.4,0.4\}$, $(B\cap C)(x_6)=\{0.1,0.1,0.3,0.3,0.4\}$. $(A\cap C)(x_6)\not\subset_m (B\cap C)(x_6)$. $(A\cap C)(x_6)\not\subset_a (B\cap C)(x_6)$. $(A\cap C)(x_6)\not\subset_t (B\cap C)(x_6)$.

$A(x_7)\subset_t B(x_7)$, $(A\cup C)(x_7)=\{0.3,0.4,0.4\}$, $(B\cup C)(x_7)=\{0.1,0.1,0.1,0.3,0.4,0.5\}$. $(A\cup C)(x_7)\not\subset_m(B\cup C)(x_7)$. $(A\cup C)(x_7)\not\subset_a(B\cup C)(x_7)$.

$A(x_6)\subset_t B(x_6)$, $(A\cup C)(x_6)\not\subset_t(B\cup C)(x_6)$.

(6) $A(x_5)\subset_n B(x_5)$, $(A\cap C)(x_5)\not\subset_m (B\cap C)(x_5)$. $(A\cap C)(x_5)\not\subset_t (B\cap C)(x_5)$.

$A(x_8)\subset_n B(x_8)$, $(A\cup C)(x_8)=\{0.8,0.9\}$, $(B\cup C)(x_8)=\{0.8,0.8,0.9\}$. $(A\cup C)(x_8)\not\subset_m (B\cup C)(x_8)$.

$A(x_5)\subset_n B(x_5)$, $(A\cup C)(x_5)\not\subset_t (B\cup C)(x_5)$.
\end{exam}

For three classical sets $\bar{A}$, $\bar{B}$ and $\bar{C}$, $\bar{A}\sqsubset \bar{B}$ and $\bar{A}\sqsubset \bar{C}$ if and only if $\bar{A}\sqsubset \bar{B}\sqcap \bar{C}$. This rule in the cases of three hesitant fuzzy sets $A$, $B$ and $C$ is investigated, shown as Proposition \ref{pro2.17}.

\begin{pro}\label{pro2.17} The following statements hold for $A,B,C\in HF(U)$,

(1) $A\subset_p B$ and $A\subset_p C$ if and only if $A\subset_p B\cap C$.

(2) $A\subset_a B$ and $A\subset_a C$ if and only if $A\subset_a B\cap C$.

(3) $A\subset_t B$ and $A\subset_t C$, then  $A\subset_t B\cap C$.

(4) $A\subset_n B$ and $A\subset_n C$ if and only if $A\subset_n B\cap C$.
\end{pro}

{\bf\slshape Proof}  (1) For all $x\in U$, $A^+(x)\leqslant B^+(x)$ and $A^+(x)\leqslant C^+(x)$ if and only if $A^+(x)\leqslant inf\{B^+(x),C^+(x)\}$.

(2) For all $x\in U$, $A^-(x)\leqslant B^-(x)$ and $A^-(x)\leqslant C^-(x)$ if and only if $A^-(x)\leqslant inf\{B^-(x),C^-(x)\}$. To combine the result of (1), then (2) holds.

(3) For a random $x\in U$, the relationships of $h\in C(x)$ and interval $[B^-(x),B^+(x)]$ are divided into two cases  (i) and (ii) as follows,

(i) $h\not\in[B^-(x),B^+(x)]$ for all $h\in C(x)$.

This case includes two sub-cases, $B^+(x)<C^-(x)$ and $C^+(x)<B^-(x)$. When $B^+(x)<C^-(x)$, $A(x)\subset_t B(x)=(B\cap C)(x)$. When $C^+(x)<B^-(x)$, $A(x)\subset_t C(x)=(B\cap C)(x)$.

(ii) $h\in[B^-(x),B^+(x)]$ for some $h\in C(x)$.

This case includes two sub-cases, $B^+(x)\leqslant C^+(x)$ and $C^+(x)< B^+(x)$.

When $B^+(x)\leqslant C^+(x)$, $h\in(B\cap C)(x)$ for all $h\in B(x)$, i.e., $B(x)\sqsubset(B\cap C)(x)$. Since $A(x)\subset_tB(x)$, by Remark \ref{proadd} (1), there is a subsequence of $B(x)$, denoted as $\overline{B(x)}$, satisfying $A(x)\preccurlyeq\overline{B(x)}$. Since $B(x)\sqsubset(B\cap C)(x)$, thus $\overline{B(x)}$ is a subsequence of $(B\cap C)(x)$, then $A(x)\subset_t (B\cap C)(x)$.

For another sub-case $C^+(x)< B^+(x)$, $h\in(B\cap C)(x)$ for all $h\in C(x)$, i.e., $C(x)\sqsubset(B\cap C)(x)$. Since $A(x)\subset_tC(x)$,  by Remark \ref{proadd} (1), there is a subsequence of $C(x)$, denoted as $\overline{C(x)}$, satisfying $A(x)\preccurlyeq\overline{C(x)}$. Since $C(x)\sqsubset(B\cap C)(x)$, thus $\overline{C(x)}$ is a subsequence of $(B\cap C)(x)$, then $A(x)\subset_t (B\cap C)(x)$.

To sum up, $A(x)\subset_t (B\cap C)(x)$ for random $x\in U$, then $A\subset_t B\cap C$.

(4) For all $x\in U$, $A^+(x)\leqslant B^-(x)$ and $A^+(x)\leqslant C^-(x)$ if and only if $A^+(x)\leqslant inf\{B^-(x),C^-(x)\}$.$\blacksquare$\\

Suppose $A(x)=\{0.3,0.5,0.7\}$, $B(x)=\{0.8,0.9\}$, $C(x)=\{0.6,0.8,0.9\}$. $A(x)\subset_t (B\cap C)(x)$, but $A(x)\not\subset_t B(x)$. The converse of Proposition \ref{pro2.17} (3) does not hold.

\subsection{The proposition of the complement of hesitant fuzzy set and the transitivity of the inclusion relationships in hesitant fuzzy sets}

\begin{pro} The following statements hold for $A,B\in HF(U)$,

(1) $A\subset_a B$, then $B^c\subset_a A^c$.

(2) $A\subset_m B$, then $B^c\subset_m A^c$.

(3) $A\subset_s B$, then $B^c\subset_{sot} A^c$.

(4) $A\subset_n B$, then $B^c\subset_n A^c$.
\end{pro}

{\bf\slshape Proof} (1) For all $x\in U$, since $A\subset_a B$, then $A^+(x)\leqslant B^+(x)$ and $A^-(x)\leqslant B^-(x)$. Then $(B^c)^-(x)=1-B^+(x)\leqslant 1-A^+(x)=(A^c)^-(x)$ and $(B^c)^+(x)=1-B^-(x)\leqslant 1-A^-(x)=(A^c)^+(x)$. Then $B^c\subset_a A^c$.

(2) For each $x\in U$, and each $H\in HF(U)$, $mean[H^c(x)]=(|H(x)|\times1-|H(x)|\times mean[H(x)])/|H^c(x)|$, since $|H(x)|=|H^c(x)|$, then $mean[H^c(x)]=1-mean[H(x)]$. $A\subset_m B$, thus $mean[A(x)]\leqslant mean[B(x)]$, then $1-mean[B(x)]\leqslant 1-mean[A(x)]$, it implies $B^c(x)\subset_m A^c(x)$.

(3) For each $x\in U$, we can sort the elements of $A(x)$ and $B(x)$ at first. Let $A(x)=\{h_1,h_2,\cdots,h_l\}$ and  $B(x)=\{h'_1,h'_2,\cdots,h'_k\}$ be two descending sequences.

Since $A\subset_s B$, then $h'_1\geqslant h_1$, $h'_2\geqslant h_2$, $\cdots$, $h'_k\geqslant h_k$ and $k\leqslant l$. Then, $1-h'_1\leqslant1-h_1$, $1-h'_2\leqslant1-h_2$, $\cdots$, $1-h'_k\leqslant1-h_k$. There is a subsequence of $A^c(x)$, denoted as $\overline{A^c(x)}$, satisfying $B^c(x)\preccurlyeq \overline{A^c(x)}$.

Since $|A^c(x)|=|A(x)|\geqslant|B(x)|=|B^c(x)|$, by Remark \ref{proadd} (1) and (2),  then $B^c(x)\subset_{sot} A^c(x)$.

(4) For all $x\in U$, since $A\subset_n B$, then $A^+(x)\leqslant B^-(x)$. $(A^c)^-(x)=1-A^+(x)\geqslant 1-B^-(x)=(B^c)^+(x)$, then $B^c(x)\subset_n A^c(x)$.$\blacksquare$

\begin{exam} Let $U=\{x,y\}$, $A=\frac{\{0.4,0.4\}}{x}+\frac{\{0.2,0.25\}}{y}$, $B=\frac{\{0.1,0.1,0.41\}}{x}+\frac{\{0.1,0.2,0.3\}}{y}$.

(1) $A(x)\subset_p B(x)$. $A^c(x)=\{0.6,0.6\}$, $B^c(x)=\{0.9,0.9,0.59\}$.  $B^c(x)\not\subset_p A^c(x)$, $B^c(x)\not\subset_m A^c(x)$.

(2) $A(y)\subset_t B(y)$. $A^c(y)=\{0.8,0.75\}$, $B^c(y)=\{0.9,0.8,0.7\}$.  $B^c(y)\not\subset_p A^c(y)$, $B^c(y)\not\subset_a A^c(y)$, $B^c(y)\not\subset_m A^c(y)$, $B^c(y)\not\subset_s A^c(y)$, $B^c(y)\not\subset_t A^c(y)$, $B^c(y)\not\subset_n A^c(y)$.
\end{exam}

It should be noted that $$A\subset_p B\nRightarrow B^c\subset_p A^c.$$

\begin{pro}\label{pro2.7} The following statements hold for $A,B,C\in HF(U)$,

(1) If $A\subset_p B$ and $B\subset_p C$, then $A\subset_p C$.

(2) If $A\subset_a B$ and $B\subset_a C$, then $A\subset_a C$.

(3) If $A\subset_m B$ and $B\subset_m C$, then $A\subset_m C$.

(4) If $A\subset_s B$ and $B\subset_s C$, then $A\subset_s C$.

(5) If $A\subset_t B$ and $B\subset_t C$, then $A\subset_t C$.

(6) If $A\subset_n B$ and $B\subset_n C$, then $A\subset_n C$.

\end{pro}

{\bf\slshape Proof} (1) and (2) are obvious.

(3) $A\subset_m B$ and $B\subset_m C$ imply that $mean[A(x)]\leqslant mean[B(x)]\leqslant mean[C(x)]$ for all $x\in U$, where $mean[\cdot]$ is the mean value operator. Then $A\subset_m C$.

(4) $A\subset_s B$ and $B\subset_s C$ imply that $|A(x)|\geqslant|B(x)|\geqslant|C(x)|$ for all $x\in U$.

Let $C(x)=\{h^C_1,h^C_2,\cdots,h^C_{|C(x)|}\}$, $B(x)=\{h^B_1,h^B_2,\cdots,h^B_{|B(x)|}\}$ and $A(x)=\{h^A_1,h^A_2,\cdots,h^A_{|A(x)|}\}$ be three descending sequences, since $A(x)\subset_s B(x)$ and $B(x)\subset_s C(x)$, then  $h^C_1\geqslant h^B_1\geqslant h^A_1$, $h^C_2\geqslant h^B_2\geqslant h^A_2$,$\cdots$, and $h^C_{|C(x)|}\geqslant h^B_{|C(x)|}\geqslant h^A_{|C(x)|}$. Hence, $A\subset_s C$.

(5) $A\subset_t B$ and $B\subset_t C$ imply that $|A(x)|<|B(x)|<|C(x)|$ for all $x\in U$.

Let $C(x)=\{h^C_1,h^C_2,\cdots,h^C_{|C(x)|}\}$, $B(x)=\{h^B_1,h^B_2,\cdots,h^B_{|B(x)|}\}$ and $A(x)=\{h^A_1,h^A_2,\cdots,h^A_{|A(x)|}\}$ be three descending sequences, since $A(x)\subset_t B(x)$ and $B(x)\subset_t C(x)$, then  $h^C_1\geqslant h^B_1\geqslant h^A_1$, $h^C_2\geqslant h^B_2\geqslant h^A_2$,$\cdots$, and $h^C_{|A(x)|}\geqslant h^B_{|A(x)|}\geqslant h^A_{|A(x)|}$. Hence, $A\subset_t C$.

(6) $A\subset_n B$ and $B\subset_n C$ imply that $A^+(x)\leqslant B^-(x)\leqslant B^+(x)\leqslant C^-(x)$ for all $x\in U$, then $A\subset_n C$.$\blacksquare$

\subsection{The situation of equality in hesitant fuzzy sets}

\begin{thm}\label{thm1} The following statements hold for $x\in U$ and $A,B\in HF(U)$,

(1) $A= B$, then $A=_p B$, $A=_a B$ and $A=_m B$.

(2) $A=_s B$ $\Leftrightarrow$ $A= B$.

(3) $A=_s B$, then $A=_p B$, $A=_a B$ and $A=_m B$.

(4) If $A=_n B$, then $h'=h''$ for all $h'\in A(x)$ and all $h''\in B(x)$.

(5) If $A=_n B$, then $A=_p B$, $A=_a B$ and $A=_m B$.
\end{thm}

{\bf\slshape Proof}  (1) Obvious.

(2) It is obvious that $A= B$ implies $A=_s B$.

On the other hand, for each $x\in U$, we can sort $A(x)$ and $B(x)$ to obtain two descending sequences $A(x)=\{h_1,h_2,\cdots,h_l\}$ and $B(x)=\{g_1,g_2,\cdots,g_k\}$. Since $A=_s B$, i.e., $A(x)\subset_s B(x)$ and $B(x)\subset_s A(x)$. Then $|A(x)|=|B(x)|$, $h_i\leqslant g_i$ and $g_i\leqslant h_i$ for $1\leqslant i\leqslant |A(x)|$. It means that $A= B$.

(3) By (2) and (1), then (3) is obvious.

(4) For each $x\in U$, if $A=_n B$, i.e., $A(x)\subset_n B(x)$ and $B(x)\subset_n A(x)$, then $B^+(x)\geqslant B^-(x)\geqslant A^+(x)\geqslant A^-(x)\geqslant B^+(x)$, then $h'=h''$ for all $h'\in A(x)$ and all $h''\in B(x)$.

(5) By (4), (5) is obvious.$\blacksquare$\\

Theorem \ref{thm1} indicates that if one of $A=_p B$, $A=_a B$ and $A=_m B$ does not hold then $A=_n B$ and $A= B$ ($A=_s B$) do not hold.

\begin{thm}\label{thm6} The following statements hold for $A,B,C\in HF(U)$,

(1) $A\cap A=_p A$, $A\cup A=_p A$.

(2) $A\cap A=_a A$, $A\cup A=_a A$.

(3) $A\cap A=_m A$, $A\cup A=_m A$.

(4) $(A\cup B)\cap A=_p A$, $(A\cap B)\cup A=_p A$.

(5) $(A\cup B)\cap A=_a A$, $(A\cap B)\cup A=_a A$.

(6) $(A\cup B)\cap C=_p (C\cap A)\cup(C\cap B)$, $(A\cap B)\cup C=_p (C\cup A)\cap(C\cup B)$.

(7) $(A\cup B)\cap C=_a (C\cap A)\cup(C\cap B)$, $(A\cap B)\cup C=_a (C\cup A)\cap(C\cup B)$.
\end{thm}

{\bf\slshape Proof}  (1)-(3) are obvious.

Let $A'$, $A''$, $B'$, $B''$, $C'$ and $C''$ be six fuzzy sets. For all $x\in U$, let $A'(x)=A^+(x)$, $A''(x)=A^-(x)$, $B'(x)=B^+(x)$, $B''(x)=B^-(x)$, $C'(x)=C^+(x)$ and $C''(x)=C^-(x)$.

For the proofs of (4)-(7), we give a simple example as follows. By the results of fuzzy sets \cite{ZADE}, $(A'(x)\cup B'(x))\cap A'(x)=A'(x)$ and $(A''(x)\cup B''(x))\cap A''(x)=A''(x)$, i.e., $(A^+(x)\cup B^+(x))\cap A^+(x)=A^+(x)$ and $(A^-(x)\cup B^-(x))\cap A^-(x)=A^-(x)$, then $(A\cup B)\cap A=_a A$.

(4)-(7) are easy to be proved by the results of fuzzy sets in  \cite{ZADE}.$\blacksquare$

\begin{exam}\label{exam1} Let $U=\{x\}$, $A=\frac{\{0.1,0.2,0.3\}}{x}$, $B=\frac{\{0.3,0.4,0.5\}}{x}$, $C=\frac{\{0.3,0.45,0.5\}}{x}$.

$(A\cap B)(x)=\{0.1,0.2,0.3,0.3\}$, $((A\cap B)\cup A)(x)=\{0.1,0,1,0.2,0.2,0.3,0.3,0.3\}\neq_m A(x)$.

$(A\cup B)(x)=\{0.3,0.3,0.4,0.5\}$, $((A\cup B)\cap A)(x)=\{0.1,0.2,0.3,0.3,0.3\}\neq_m A(x)$.

$((A\cup B)\cap C)(x)=\{0.3,0.3,0.3,0.4,0.45,0.5\}$, $((A\cap C)\cup(B\cap C))(x)=\{0.3,0.3,0.3,0.3,0.4,0.45,0.5\}$, $((A\cup B)\cap C)(x)\neq_m((A\cap C)\cup(B\cap C))(x)$.

$((A\cap B)\cup C)(x)=\{0.3,0.3,0.3,0.45,0.5\}$, $((A\cup C)\cap(B\cup C))(x)=\{0.3,0.3,0.3,0.3,0.4,0.45,0.45,0.5,0.5\}$, $((A\cap B)\cup C)(x)\neq_m((A\cup C)\cap(B\cup C))(x)$.

In addition, $0.4\not\in((A\cap B)\cup C)(x)$, however, $0.4\in((A\cup C)\cap(B\cup C))(x)$.
\end{exam}

Inspired by Theorem \ref{thm1} and Example \ref{exam1}, it should be noted that
$$\begin{cases}
(A\cap B)\cup A\not= A,\\
(A\cup B)\cap A\not= A,\\
(A\cup B)\cap C\not=(A\cap C)\cup(B\cap C),\\
(A\cap B)\cup C\not=(A\cup C)\cap(B\cup C),\\
(A\cap B)\cup A\neq_n A,\\
(A\cup B)\cap A\neq_n A,\\
(A\cup B)\cap C\neq_n(A\cap C)\cup(B\cap C),\\
(A\cap B)\cup C\neq_n(A\cup C)\cap(B\cup C).\\
\end{cases}$$

When considering the $\subset_p$ and $=_p$  relationships, the primary focus lies on determining the upper bound for an object $x\in U$ and a hesitant fuzzy set defined over $U$.
Consequently, several inclusion and equal situations that are applicable to fuzzy sets can also be extended to hesitant fuzzy sets in the form of $\subset_p$ and $=_p$ relationships.
Nevertheless, the absence of the implication $A\subset_p B\nRightarrow B^c\subset_p A^c$ indicates that the $\subset_p$ and $=_p$ relationships do not align with the inclusion relationships and equal situations observed in fuzzy sets.

Sections 2.2 and 2.3 reveal that certain commonly observed inclusion  and equal situations do not apply to hesitant fuzzy sets but are valid for classical sets.

\section{Foundations of families of hesitant fuzzy sets}

Let $\mathscr{H}_1$ and $\mathscr{H}_2$ be two families of hesitant fuzzy sets. If $H_i\in \mathscr{H}_2$ holds for all $H_i\in \mathscr{H}_1$, we say that $\mathscr{H}_1$ is the classical subset of $\mathscr{H}_2$, denoted as $\mathscr{H}_1\sqsubset \mathscr{H}_2$.

\begin{thm}\label{thm2} Let  $\mathscr{H}\sqsubset HF(U)$ be a family of  hesitant fuzzy sets over $U$.  The following statements hold,

(1) $\bigcap\{H: H\in \mathscr{H}\}\subset_p\bigcup\{H: H\in \mathscr{H}\}$.

(2) $\bigcap\{H: H\in \mathscr{H}\}\subset_a\bigcup\{H: H\in \mathscr{H}\}$.

(3) $\bigcap\{H: H\in \mathscr{H}\}\subset_m\bigcup\{H: H\in \mathscr{H}\}$.

(4) $\bigcap\{H: H\in \mathscr{H}\}\subset_{sot}\bigcup\{H: H\in \mathscr{H}\}$.

\end{thm}

{\bf\slshape Proof} (1) For each $x\in U$, $inf\{H^+(x): H\in \mathscr{H}\}\leqslant sup\{H^+(x): H\in \mathscr{H}\}$.

(2) For each $x\in U$, $inf\{H^-(x): H\in \mathscr{H}\}\leqslant sup\{H^-(x): H\in \mathscr{H}\}$. To combine the result of (1), then (2) holds.

(3) For each $x\in U$, let $H_{inf}^+(x)=inf\{H^+(x): H\in \mathscr{H}\}$, $H_{sup}^-(x)=sup\{H^-(x): H\in \mathscr{H}\}$.

If $H_{inf}^+(x)\leqslant H_{sup}^-(x)$, $(\bigcap\{H: H\in \mathscr{H}\})(x)\subset_m(\bigcup\{H: H\in \mathscr{H}\})(x)$ is obvious.

If $H_{sup}^-(x)<H_{inf}^+(x)$, we construct a set of numbers $V=\{h:h\in H(x), H\in \mathscr{H},H_{sup}^-(x)\leqslant h\leqslant H_{inf}^+(x)\}$.

If $(\bigcap\{H: H\in \mathscr{H}\})(x)-V\neq\emptyset$, then $w<inf(V)$ for all $w\in (\bigcap\{H: H\in \mathscr{H}\})(x)-V$. If $(\bigcup\{H: H\in \mathscr{H}\})(x)-V\neq\emptyset$, then $w'>sup(V)$ for all $w'\in(\bigcup\{H: H\in \mathscr{H}\})(x)-V$. Then, $mean[(\bigcap\{H: H\in \mathscr{H}\})(x)]\leqslant mean[V]\leqslant mean[(\bigcup\{H: H\in \mathscr{H}\})(x)]$, i.e., $(\bigcap\{H: H\in \mathscr{H}\})(x)\subset_m(\bigcup\{H: H\in \mathscr{H}\})(x)$.

(4) For each $x\in U$, $H_{inf}^+(x)$ and $H_{sup}^-(x)$ are constructed as (2) above.

If $H_{inf}^+(x)\leqslant H_{sup}^-(x)$, by Proposition \ref{pro9ch} (9), $(\bigcap\{H: H\in \mathscr{H}\})(x)\subset_{sot}(\bigcup\{H: H\in \mathscr{H}\})(x)$ is obvious.

If $H_{sup}^-(x)<H_{inf}^+(x)$, let $V=(\bigcap\{H: H\in \mathscr{H}\})(x)\sqcap(\bigcup\{H: H\in \mathscr{H}\})(x)$, $inf((\bigcup\{H: H\in \mathscr{H}\})(x)-V)>H_{inf}^+(x)>H_{sup}^-(x)>sup((\bigcap\{H: H\in \mathscr{H}\})(x)-V)$, by Remark \ref{proadd} (4),
$(\bigcap\{H: H\in \mathscr{H}\})(x)\subset_{sot} (\bigcup\{H: H\in \mathscr{H}\})(x)$.$\blacksquare$

\begin{thm}\label{thm2.26}  Let $A\in HF(U)$ and $\mathscr{H}\sqsubset HF(U)$ be a family of  hesitant fuzzy sets over $U$. The following statements hold,

(1) $A\subset_p H$ for all $H\in\mathscr{H}$ if and only if $A\subset_p\bigcap\{H: H\in \mathscr{H}\}$.

(2) If $A\subset_p H_{\alpha}$ for a $H_{\alpha}\in\mathscr{H}$, then $A\subset_p\bigcup\{H: H\in \mathscr{H}\}$.

(3) $A\subset_a H$ for all $H\in\mathscr{H}$ if and only if $A\subset_a\bigcap\{H: H\in \mathscr{H}\}$.

(4) If $A\subset_a H_{\alpha}$ for a $H_{\alpha}\in\mathscr{H}$, then $A\subset_a\bigcup\{H: H\in \mathscr{H}\}$.

(5) If $A\subset_t H$ for all $H\in\mathscr{H}$, then $A\subset_t\bigcap\{H: H\in \mathscr{H}\}$.

(6) If $A\subset_t H_{\alpha}$ for a $H_{\alpha}\in\mathscr{H}$, and $|A(x)|<|H(x)|$ for all $H\in\mathscr{H}$ and $x\in U$, then $A\subset_t\bigcup\{H: H\in \mathscr{H}\}$.

(7) $A\subset_n H$ for all $H\in\mathscr{H}$ if and only if $A\subset_n\bigcap\{H: H\in \mathscr{H}\}$.

(8) If $A\subset_n H_{\alpha}$ for a $H_{\alpha}\in\mathscr{H}$, then $A\subset_n\bigcup\{H: H\in \mathscr{H}\}$.
\end{thm}

{\bf\slshape Proof} (1) For each $x\in U$, $A^+(x)\leqslant inf\{H^+(x):H\in\mathscr{H}\}$ if and only if $A^+(x)\leqslant H^+(x)$ for all $H\in\mathscr{H}$.

(2) $A\subset_pH_{\alpha}\in\mathscr{H}$, $A^+(x)\leqslant H_{\alpha}^+(x)\leqslant sup\{H^+(x):H\in \mathscr{H}\}$ for all $x\in U$.

(3) For each $x\in U$, $A^-(x)\leqslant inf\{H^-(x):H\in\mathscr{H}\}$ if and only if $A^-(x)\leqslant H^-(x)$ for all $H\in\mathscr{H}$. To combine the result of (1), then (3) holds.

(4) $A\subset_aH_{\alpha}\in\mathscr{H}$, $A^+(x)\leqslant H_{\alpha}^+(x)\leqslant sup\{H^+(x):H\in \mathscr{H}\}$ and $A^-(x)\leqslant H_{\alpha}^-(x)\leqslant sup\{H^-(x):H\in \mathscr{H}\}$ for all $x\in U$.

(5) For each $x\in U$, $A\subset_t H$ for all $H\in\mathscr{H}$, then $|A(x)|<|H(x)|$ and $|A(x)|<|(\bigcap\{H: H\in \mathscr{H}\})(x)|$ are obvious.

$H_{\gamma}(x)$ can be sorted into a descending sequence $H_{\gamma}(x)=\{h_{\gamma}^1,h_{\gamma}^2,\cdots\}$ for each $H_{\gamma}\in\mathscr{H}$ and $\gamma\in\Gamma$, where $\Gamma$ is an order set and $|\Gamma|=|\mathscr{H}|$.

Let $\overline{H}=\{\overline{h}^1,\overline{h}^2,\cdots,\overline{h}^{|A(x)|}\}$ be a subsequence of $(\bigcap\{H_{\gamma}: H_{\gamma}\in \mathscr{H},{\gamma}\in\Gamma\})(x)$, where $\overline{h}^j=inf\{h_{\gamma}^j:h_{\gamma}^j\in H_{\gamma}(x),\gamma\in\Gamma\}$ for $1\leqslant j\leqslant|A(x)|$.

$A(x)$ can be sorted into a descending sequence $A(x)=\{v_1,v_2,\cdots,v_{|A(x)|}\}$. Since $h_{\gamma}^j\geqslant v_j$ for all $\gamma\in\Gamma$ and $1\leqslant j\leqslant|A(x)|$, then $\overline{h}^j=inf\{h_{\gamma}^j:h_{\gamma}^j\in H_{\gamma}(x),\gamma\in\Gamma\}\geqslant v_j$ for $1\leqslant j\leqslant|A(x)|$, i.e., $A(x)\preccurlyeq \overline{H}$.

$|A(x)|<|(\bigcap\{H: H\in \mathscr{H}\})(x)|$, by Remark \ref{proadd} (1), then $A(x)\subset_t(\bigcap\{H: H\in \mathscr{H}\})(x)$.

(6) $|A(x)|<|H(x)|$ for all $H\in \mathscr{H}$ and $x\in U$. Let $H_*=\{h_*^1,h_*^2,\cdots,h_*^{|A(x)|}\}$ be a subsequence of $(\bigcup\{H_{\gamma}: H_{\gamma}\in \mathscr{H},{\gamma}\in\Gamma\})(x)$, where $h_*^j=sup\{h_{\gamma}^j:h_{\gamma}^j\in H_{\gamma}(x),\gamma\in\Gamma\}$ for $1\leqslant j\leqslant|A(x)|$.

For two descending sequences $A(x)=\{v_1,v_2,\cdots,v_{|A(x)|}\}$ and $H_{\alpha}(x)=\{h_{\alpha}^1,h_{\alpha}^2,\cdots\}$, since $A\subset_tH_{\alpha}$, then $v_j\leqslant h_{\alpha}^j\leqslant h_*^j=sup\{h_{\gamma}^j:h_{\gamma}^j\in H_{\gamma}(x),\gamma\in\Gamma\}$ for $1\leqslant j\leqslant|A(x)|$, i.e., $A(x)\preccurlyeq H_*$.

$|A(x)|<|(\bigcup\{H: H\in \mathscr{H}\})(x)|$, by Remark \ref{proadd} (1), then $A(x)\subset_t(\bigcup\{H: H\in \mathscr{H}\})(x)$.

(7) For each $x\in U$, $A^+(x)\leqslant inf\{H^-(x):H\in \mathscr{H}\}$ if and only if $A^+(x)\leqslant H^-(x)$ for all $H\in \mathscr{H}$.

(8) $A\subset_nH_{\alpha}\in \mathscr{H}$, then $A^+(x)\leqslant H_{\alpha}^-(x)\leqslant sup\{H^-(x):H\in \mathscr{H}\}$ for all $x\in U$.$\blacksquare$

\begin{thm}\label{thm3} Let $\mathscr{H}_1,\mathscr{H}_2\sqsubset HF(U)$ be two families of  hesitant fuzzy sets over $U$. The following statements hold,

(1) If $\mathscr{H}_1\sqsubset\mathscr{H}_2$, then $\bigcap\{H: H\in \mathscr{H}_2\}\subset_p\bigcap\{H: H\in \mathscr{H}_1\}$.

(2) If $\mathscr{H}_1\sqsubset\mathscr{H}_2$, then $\bigcup\{H: H\in \mathscr{H}_1\}\subset_p\bigcup\{H: H\in \mathscr{H}_2\}$.

(3)  If $\mathscr{H}_1\sqsubset\mathscr{H}_2$, then $\bigcap\{H: H\in \mathscr{H}_1\}\subset_p\bigcup\{H: H\in \mathscr{H}_2\}$.

(4) If $\mathscr{H}_1\sqsubset\mathscr{H}_2$, then $\bigcap\{H: H\in \mathscr{H}_2\}\subset_a\bigcap\{H: H\in \mathscr{H}_1\}$.

(5) If $\mathscr{H}_1\sqsubset\mathscr{H}_2$, then $\bigcup\{H: H\in \mathscr{H}_1\}\subset_a\bigcup\{H: H\in \mathscr{H}_2\}$.

(6)  If $\mathscr{H}_1\sqsubset\mathscr{H}_2$, then $\bigcap\{H: H\in \mathscr{H}_1\}\subset_a\bigcup\{H: H\in \mathscr{H}_2\}$.

(7) If $\mathscr{H}_1\sqsubset\mathscr{H}_2$, then $\bigcup\{H: H\in \mathscr{H}_1\}\subset_{sot}\bigcup\{H: H\in \mathscr{H}_2\}$.

(8)  If $\mathscr{H}_1\sqsubset\mathscr{H}_2$, then $\bigcap\{H: H\in \mathscr{H}_1\}\subset_{sot}\bigcup\{H: H\in \mathscr{H}_2\}$.
\end{thm}

{\bf\slshape Proof} (1) For each $x\in U$, if $\mathscr{H}_1\sqsubset\mathscr{H}_2$, then $\{H^+(x):H\in\mathscr{H}_1\}\sqsubset\{H^+(x):H\in\mathscr{H}_2\}$. $inf\{H^+(x):H\in\mathscr{H}_1\}\geqslant inf\{H^+(x):H\in\mathscr{H}_2\}$, then $(\bigcap\{H: H\in \mathscr{H}_2\})(x)\subset_p(\bigcap\{H: H\in \mathscr{H}_1\})(x)$.

(2) For each $x\in U$, if $\mathscr{H}_1\sqsubset\mathscr{H}_2$, then $\{H^+(x):H\in\mathscr{H}_1\}\sqsubset\{H^+(x):H\in\mathscr{H}_2\}$. $sup\{H^+(x):H\in\mathscr{H}_1\}\leqslant sup\{H^+(x):H\in\mathscr{H}_2\}$, then $(\bigcup\{H: H\in \mathscr{H}_1\})(x)\subset_p(\bigcup\{H: H\in \mathscr{H}_2\})(x)$.

(3) If $\mathscr{H}_1\sqsubset\mathscr{H}_2$, by Theorem \ref{thm2} (1) and Theorem \ref{thm3} (2), $\bigcap\{H: H\in \mathscr{H}_1\}\subset_p\bigcup\{H: H\in \mathscr{H}_1\}\subset_p\bigcup\{H: H\in \mathscr{H}_2\}$.

(4) For each $x\in U$, if $\mathscr{H}_1\sqsubset\mathscr{H}_2$, then $\{H^-(x):H\in\mathscr{H}_1\}\sqsubset\{H^-(x):H\in\mathscr{H}_2\}$. $inf\{H^-(x):H\in\mathscr{H}_1\}\geqslant inf\{H^-(x):H\in\mathscr{H}_2\}$. To combine the result of (1), then (4) holds.

(5) For each $x\in U$, if $\mathscr{H}_1\sqsubset\mathscr{H}_2$, then $\{H^-(x):H\in\mathscr{H}_1\}\sqsubset\{H^-(x):H\in\mathscr{H}_2\}$. $sup\{H^-(x):H\in\mathscr{H}_1\}\leqslant sup\{H^-(x):H\in\mathscr{H}_2\}$. To combine the result of (2), then (5) holds.

(6) If $\mathscr{H}_1\sqsubset\mathscr{H}_2$, by Theorem \ref{thm2} (2) and Theorem \ref{thm3} (5), $\bigcap\{H: H\in \mathscr{H}_1\}\subset_a\bigcup\{H: H\in \mathscr{H}_1\}\subset_a\bigcup\{H: H\in \mathscr{H}_2\}$.

(7) For each $x\in U$, let $sup_1^-=sup\{H^-(x):H\in\mathscr{H}_1\}=(\bigcup\{H: H\in \mathscr{H}_1\})^-(x)$, $sup_2^-=sup\{H^-(x):H\in\mathscr{H}_2\}=(\bigcup\{H: H\in \mathscr{H}_2\})^-(x)$.

Let $V=(\bigcup\{H: H\in \mathscr{H}_1\})(x)\sqcap (\bigcup\{H: H\in \mathscr{H}_2\})(x)$. Since $\mathscr{H}_1\sqsubset\mathscr{H}_2$, then $sup_1^-\leqslant sup_2^-$. If $h\in(\bigcup\{H: H\in \mathscr{H}_1\})(x)$ and $sup_2^-\leqslant h$, then $h\in(\bigcup\{H: H\in \mathscr{H}_2\})(x)$. It means that $sup((\bigcup\{H: H\in \mathscr{H}_1\})(x)-V)<sup_2^-$. On the other hand, $sup_2^-<inf((\bigcup\{H: H\in \mathscr{H}_2\})(x)-V)$, by Remark \ref{proadd} (4), then $(\bigcup\{H: H\in \mathscr{H}_1\})(x)\subset_{sot} (\bigcup\{H: H\in \mathscr{H}_2\})(x)$.

(8) For each $x\in U$, let $inf_1^+=inf\{H^+(x): H\in \mathscr{H}_1\}$, $sup_2^-=sup\{H^-(x): H\in \mathscr{H}_2\}$.

If $inf_1^+\leqslant sup_2^-$, by Proposition \ref{pro9ch}
 (9), $(\bigcap\{H: H\in \mathscr{H}_1\})(x)\subset_{sot}(\bigcup\{H: H\in \mathscr{H}_2\})(x)$ is obvious.

If $sup_2^-<inf_1^+$. Let $V=(\bigcap\{H: H\in \mathscr{H}_1\})(x)\sqcap (\bigcup\{H: H\in \mathscr{H}_2\})(x)$.

Base on the condition $\mathscr{H}_1\sqsubset\mathscr{H}_2$, if $h\in(\bigcap\{H: H\in \mathscr{H}_1\})(x)$ and $sup_2^-\leqslant h$, then $h\in(\bigcup\{H: H\in \mathscr{H}_2\})(x)$. It means that $sup((\bigcap\{H: H\in \mathscr{H}_1\})(x)-V)<sup_2^-$. On the other hand, $sup_2^-<inf((\bigcup\{H: H\in \mathscr{H}_2\})(x)-V)$, by Remark \ref{proadd} (4), then $(\bigcap\{H: H\in \mathscr{H}_1\})(x)\subset_{sot} (\bigcup\{H: H\in \mathscr{H}_2\})(x)$.$\blacksquare$

\section{Conclusion}

The definition of the inclusion relationship is one of the most foundational concepts of sets.
Based on the discrete form of hesitant fuzzy membership degrees,
we propose several kinds of inclusion relationships for hesitant fuzzy sets.   Furthermore, we present some propositions of hesitant fuzzy sets and some propositions of the families of hesitant fuzzy sets based on the proposed inclusion relationships.
Some rules that hold in classical sets do not apply to hesitant fuzzy sets, as shown in Section 2. In subsequent studies of hesitant fuzzy sets, researchers must avoid the intuitive understanding of classical sets and cannot use these rules.

In our future work, we will investigate more applications of  multi-strength intelligent classifiers, saturation reductions in HFISs, and foundational theories of knowledge bases with varying strengths of knowledge, among other topics.

\section*{Acknowledgments}

The authors thank the editors and the anonymous reviewers for
their helpful comments and suggestions that have led to this improved
version of the paper. This work was supported by the National Social Science Foundation Youth Project of China (No. 24CTQ029).

\bibliographystyle{elsarticle-num}
%\bibliographystyle{plain}
%\bibliography{biblio}
\bibliography{reference}

%\section*{References}

%\begin{thebibliography}{9}

%\footnotesize

\end{multicols}

\end{document}